\pdfoutput=1
\documentclass[]{IEEEtran}

\usepackage[noadjust]{cite}
\usepackage[colorlinks=false,linkcolor=black,urlcolor=black,bookmarksopen=true,hidelinks,hypertexnames=false]{hyperref}
\usepackage{algorithm}
\usepackage{colortbl}
\usepackage{float}

\usepackage{algpseudocode}

\usepackage{graphicx}
\usepackage{url}

\usepackage{indentfirst}
\usepackage{adjustbox}
\usepackage{bm} 
\usepackage{color} 
\usepackage{array}
\usepackage{balance}
\usepackage{marginnote} 
\usepackage{times}
\usepackage{booktabs}
\usepackage{multirow}
\usepackage{comment}

\usepackage{tabularx,booktabs,multirow}

\DeclareMathAlphabet\bmcal{OMS}{cmsy}{b}{n}

\usepackage{amsmath,amssymb,amsfonts}
\usepackage{authblk}
\usepackage{graphicx}
\usepackage{blkarray, bigstrut}
\usepackage{xspace}
\usepackage{xparse}
\usepackage{xcolor}
\usepackage[flushleft]{threeparttable}
\newcolumntype{R}[1]{>{\raggedleft\let\newline\\\arraybackslash\hspace{0pt}}m{#1}}
\usepackage[caption=false,font=footnotesize]{subfig}

\newcommand{\rev}[1]{\textcolor{black}{#1}}
\begin{document}
\raggedbottom
\hbadness=10000
\vbadness=10000
\hfuzz=1000pt
\vfuzz=1000pt




\title{ResLearn-XR: Residual Learning for Network Traffic and  Quality-of-Experience-Aware Modeling in Extended Reality}

\author{
    Yoga Suhas Kuruba Manjunath, \textit{Member, IEEE},\\ 
    Jie Gao, \textit{Senior Member, IEEE},
    Lian Zhao, \textit{Fellow, IEEE}
    
    \thanks{
    Y. Manjunath is with the School of Information Technology, Carleton University, Ottawa, ON, K1S 5B6, Canada, and the Department of Electrical, Computer and Biomedical Engineering, Toronto Metropolitan University, Toronto, ON M5B 2K3, Canada. (e-mail: yogasuhaskurubamanju@carleton.ca, yoga.kuruba@torontomu.ca).
    }
    \thanks{
    J. Gao is with the School of Information Technology, Carleton University, Ottawa, ON, K1S 5B6, Canada. (e-mail: jie.gao6@carleton.ca).
    }
    \thanks{L. Zhao is with the Department of Electrical, Computer and Biomedical Engineering, Toronto Metropolitan University, Toronto, ON M5B 2K3, Canada. (Corresponding author e-mail: l5zhao@torontomu.ca)
    }

}


\maketitle

\begin{abstract}
We present ResLearn-XR, a residual learning framework for predicting eXtended Reality (XR) network traffic and estimating Quality-of-Experience (QoE) risk. ResLearn-XR adopts a two-stage temporal learning structure comprising a base sequence prediction model augmented with task-specific residual learning components to improve adaptability to bursty, non-stationary XR traffic dynamics. The residual learning stages operate in the value space for continuous XR traffic forecasting and in the logit space for probabilistic QoE risk estimation. \rev{For the QoE-risk branch, we introduce a Data Descriptor Algorithm (DDA), a causal feature-construction module that converts packet-level application-layer observables into frame-timing-aware descriptors suitable for encrypted traffic analysis. We also construct an XR Traffic-QoE dataset that pairs continuous XR traffic traces with session-level user-reported QoE labels. ResLearn-XR reduces SMAPE by up to 17.84\% across frame-count, frame-size, and inter-arrival-time prediction, while reducing QoE-risk estimation SMAPE by up to 87.8\% over single-stage baselines.}
\end{abstract}

\IEEEpeerreviewmaketitle

\section{Introduction}
\label{sec:int}

Extended Reality (XR), encompassing Virtual Reality (VR), Augmented Reality (AR), and Mixed Reality (MR), represents one of the most demanding use cases for cognitive next-generation communication networks~\cite{9999425}. XR applications generate synchronized video, audio, and control traffic while operating under a strict motion-to-photon (MTP) latency constraint of under 20~ms~\cite{8395443}. Violations of this constraint trigger cybersickness, a sensory mismatch causing motion discomfort and severe Quality-of-Experience (QoE) degradation~\cite{8329628,10926847}. Supporting XR over wireless networks, therefore, requires communication systems capable of cognitive behaviour: continuously observing traffic dynamics, inferring imminent QoE risk, and enabling proactive resource management before MTP latency thresholds are breached.

This cognitive intelligence requirement motivates two tightly coupled learning tasks. First, XR traffic prediction must capture both long-range temporal trends and the short-term, bursty frame-level fluctuations that directly drive network queueing delay and MTP violations. Second, QoE risk estimation must map observed network behaviour to the likelihood of elevated cybersickness risk, enabling predictive, rather than reactive, network control.

Conventional temporal predictors, including time-series forecasters~\cite{lim2021temporal}, Long Short-Term Memory (LSTM) networks~\cite{10437897}, Gated Recurrent Unit (GRU) networks~\cite{woo2024improving}, and Transformer-based architectures such as Informer and FEDformer~\cite{zhou2021informer,zhou2022fedformer}, are trained with sequence-level regression objectives that bias predictions toward conditional mean behaviour. Consequently, high-variance frame-level traffic bursts are underpredicted, precisely at the time-scales critical for meeting MTP requirements. A single-stage predictor is thus insufficient to jointly model long-range trends and burst dynamics, motivating explicit residual learning for burst-phase correction.

On the QoE estimation side, the ITU-T identifies XR QoE as dependent on network performance, configuration-dependent timing targets (e.g., display refresh rate), and user-perceived experience~\cite{ITU-T2020,ITUT_P812}. Yet existing deep learning~\cite{kougioumtzidis2023deep} and digital twin~\cite{10486201} approaches rely on static or aggregated network indicators, lacking the frame-level timing-variability representations needed to capture transient QoE degradation. Furthermore, existing XR datasets~\cite{9685808,9783169,data8080132,questset} provide traffic traces without associated session-level user QoE labels, preventing supervised learning of the network traffic-to-QoE relationship.

We introduce ResLearn-XR, a two-stage residual learning framework for joint XR traffic prediction and QoE risk estimation. The framework augments a base Transformer encoder with task-specific residual heads: a value-space head corrects burst-phase traffic prediction errors, while a logit-space head refines probabilistic QoE risk estimates. \rev{DDA} transforms encrypted-traffic-compatible packet-level observables into frame-timing-aware causal descriptors aligned with ITU-T QoE causality requirements. We additionally release an XR Traffic-QoE dataset in which session-level QoE annotations are paired with the corresponding continuous traffic traces across diverse XR applications and network conditions. \rev{ResLearn-XR achieves causal SMAPE improvements of up to 17.84\% (approximately 6.7\% on average) across frame-count, frame-size, and inter-arrival-time prediction, and up to 87.8\% SMAPE improvement for QoE risk estimation.}

\rev{Compared with our prior ResLearn traffic-prediction framework~\cite{manjunath2025reslearn}, which focused on frame-level XR traffic forecasting, this paper extends the scope to joint XR traffic prediction and QoE risk estimation through logit-space residual correction and DDA-based encrypted-traffic-compatible feature construction. It further introduces the XR Traffic-QoE dataset with session-level user annotations and expands the evaluation to weakly supervised QoE risk estimation, cross-user testing, DDA ablation, calibration, latency, and proxy-based MTP-risk analysis.}

This work makes three main contributions,
\begin{itemize}
    \item \textbf{ResLearn-XR}: a two-stage Transformer framework with value-space and logit-space residual heads for XR traffic prediction and QoE risk estimation, enabling cognitive awareness of burst dynamics and perceptual risk;
    \rev{\item \textbf{DDA}: a causal feature construction method that derives frame-timing-aware causal descriptors from application-layer observables (packet length, direction, inter-arrival time, timestamp), enabling QoE risk estimation under encrypted traffic;}
    \rev{\item \textbf{XR Traffic-QoE Dataset}: a publicly released dataset covering diverse XR applications and system configurations with session-level user-reported QoE labels associated with the corresponding traffic traces, to support weakly supervised QoE risk estimation.}
\end{itemize}

To the best of our knowledge, ResLearn-XR is among the first frameworks to jointly address XR traffic prediction and QoE risk estimation in a unified cognitive architecture. The implementations for XR traffic prediction\footnote{\url{https://github.com/yoga-suhas-km/ResLearn}} and QoE risk estimation\footnote{\url{https://github.com/yoga-suhas-km/XR_QoE_Prediction}} are publicly available, as is the XR Traffic-QoE dataset\footnote{\url{https://dx.doi.org/10.21227/na8a-9n63}}.

\section{Related Work}
\label{sec:rw}

\subsection{XR Traffic Prediction}

XR traffic exhibits piecewise non-stationary behavior driven by scene dynamics, encoder rate control, and rendering synchronization~\cite{lecci2021open}, undermining statistical models such as Autoregressive Integrated Moving Average (ARIMA) and Hidden Markov Models (HMMs) that assume stationarity. Vaidya \textit{et al.} propose an LSTM-based transfer learning model for VR cloud-gaming traffic forecasting~\cite{10437897}; however, recurrent architectures remain limited in capturing abrupt frame-level bursts. Transformer-based models (Temporal Fusion Transformer~\cite{lim2021temporal}, Informer~\cite{zhou2021informer}, FEDformer~\cite{zhou2022fedformer}) improve long-range forecasting but oversmooth frame-scale bursts, underrepresenting the rapid transitions intrinsic to XR traffic, as confirmed by Mor\'{i}n \textit{et al.}~\cite{morin2023extended} and Chiariotti \textit{et al.}~\cite{chiariotti2024}. These burst phases are critical for meeting frame delivery deadlines and MTP latency requirements, motivating explicit residual modeling of short-timescale prediction errors.

\subsection{XR QoE Risk Estimation}

Ruan and Xie~\cite{ruan2021survey} identify that XR-critical timing factors such as frame-rate variability and MTP latency are only indirectly reflected in aggregated QoS metrics. Reactive approaches, cooperative rendering~\cite{liu2025qoe} and adaptive edge streaming~\cite{huang2025enhanced}, reduce latency but lack temporal learning mechanisms for predictive QoE risk estimation. Kougioumtzidis \textit{et al.}~\cite{kougioumtzidis2023deep} and Jiadong \textit{et al.}~\cite{10486201} propose LSTM-based and digital-twin-based QoE models, but neither explicitly links XR traffic evolution to frame-level delivery behavior, limiting their ability to capture how XR traffic instability translates into elevated QoE risk.

\subsection{XR Traffic-QoE Datasets}

Existing XR traffic datasets~\cite{lecci2021open,9685808,9783169,data8080132,questset} provide packet- and frame-level measurements but lack associated session-level subjective QoE labels. Studies demonstrating the sensitivity of XR experience to transient network disruptions~\cite{10740019,morin2023extended,11087572} use proprietary data. No open dataset jointly captures diverse XR traffic dynamics and corresponding session-level user QoE outcomes, limiting development of predictive XR frameworks. Table~\ref{tab:litreview_xr_traffic_qoe} contextualizes ResLearn-XR within the literature.

\begin{table*}[t]
\centering
\caption{Comparative Summary of Prior Works and Datasets in XR Traffic Prediction and QoE Risk Estimation}

\renewcommand{\arraystretch}{1.1}
\setlength{\tabcolsep}{3pt}
\begin{tabular}{
    >{\centering\arraybackslash}p{2cm}
    >{\centering\arraybackslash}p{1.5cm}
    >{\centering\arraybackslash}p{1.5cm}
    >{\centering\arraybackslash}p{1.7cm}
    >{\centering\arraybackslash}p{8cm}
}
\hline
\textbf{Reference} & \textbf{XR Traffic Prediction} & \textbf{XR QoE Risk Estimation} & \textbf{XR Traffic-QoE Data} & \textbf{Assumptions/Limitations} \\
\hline
 \cite{lecci2021open} & Yes & No & No & Assumes stationarity \\ \hline
 \cite{10437897} & Yes & No & No & Limited temporal granularity; underfits non-stationary bursts \\ \hline \cite{lim2021temporal}, \cite{zhou2021informer}, \cite{zhou2022fedformer} & Yes & No & No & Over-smooths short bursts; unsuitable for XR volatility \\ \hline
 \cite{morin2023extended}, \cite{chiariotti2024} & Yes & No & No & Lack perceptual QoE outcomes \\ \hline
 \cite{ruan2021survey} & No & Yes & No & Ignores temporal and perceptual variation in QoE \\ \hline
 \cite{liu2025qoe}, \cite{huang2025enhanced} & No & Yes & No & Frame-level only features\\ \hline
 \cite{kougioumtzidis2023deep} & No & Yes & No & Perceptual-layer only; Latency-based features only\\ \hline
 \cite{10486201} & No & Yes & No & Reactive orchestration \\ \hline
\cite{lecci2021open}, \cite{9685808,9783169,data8080132,questset} & No & No & No & No associated session-level subjective QoE labels \\ \hline
\cite{10740019}, \cite{morin2023extended} & No & No & No & Proprietary or scenario-limited \\ \hline
 \textbf{This Work} & \textbf{Yes} & \textbf{Yes} & \textbf{Yes} & - \\
\hline
\end{tabular}
\label{tab:litreview_xr_traffic_qoe}
\end{table*}

\section{Problem Formulation}
\label{sec:problem}

An XR session produces a network packet sequence $\mathbf{X} = \{\mathbf{x}_t \mid t = 1, \ldots, T\}$, where each $\mathbf{x}_t \in \mathbb{R}^{D}$ encodes packet-level attributes (direction, length, inter-arrival time, timestamp). The objective is to learn two predictive functions: (i) XR traffic prediction and (ii) QoE risk estimation, detailed below.


\medskip
\noindent
\textbf{Traffic prediction:}
We aim to learn a predictive function
\begin{equation}
    \label{eq:traffic_prediction_mapping}
    f_{\text{traf}} : \mathbb{R}^{D \times T} \rightarrow \mathbb{R}^{m},
\end{equation}
that captures the temporal evolution of XR traffic and forecasts its future state.
The function operates on an observed packet-level traffic sequence
$
\mathbf{X}
$
as defined above.

Let
$
\mathbf{y}_t = [y_t^{(1)}, \ldots, y_t^{(m)}] \in \mathbb{R}^{m}, \; m \le D,
$
denote a vector of $m$ frame-level traffic metrics, namely, frame count, average frame size,
and frame inter-arrival time, derived by aggregating the underlying packet-level observations
over a temporal window ending at time $t$. The dimensionality $m$ satisfies $m < D$ because
the traffic metrics are not independent raw features, but lower-dimensional aggregates
computed from the $D$ packet-level features. While these metrics can be computed from packet
traces after observation, they are not available at prediction time for future windows, as
their computation requires packet arrivals beyond time $T$.

Given the observed packet-level traffic sequence $\mathbf{X}$ up to time $T$, the predictor
outputs a next-step estimate
$
\widehat{\mathbf{y}}_{T+1} = f_{\text{traf}}(\mathbf{X}),
$
which aims to forecast the corresponding ground-truth traffic metrics $\mathbf{y}_{T+1}$.
The traffic prediction task learns the function $f_{\text{traf}}$ by minimizing a
task-specific loss $\mathcal{L}_{\text{traf}}(\cdot)$ between predicted and ground-truth
future traffic states.

\medskip
\noindent
\textbf{QoE Risk Estimation:}
We aim to estimate the risk of QoE violation, hereafter referred to as ``QoE risk,''
associated with an XR traffic sequence.
This task is formulated as a probabilistic mapping
\begin{equation}
\label{eq:qoe_mapping}
f_{\text{qoe}} : \mathbb{R}^{D \times T} \rightarrow [0,1],
\end{equation}
which maps an observed packet-level traffic sequence
$\mathbf{X} \in \mathbb{R}^{D \times T}$
to a scalar risk probability
$
\widehat{p} = f_{\text{qoe}}(\mathbf{X}),
$
where $\widehat{p} \in [0,1]$ represents the estimated likelihood that the traffic
sequence is associated with elevated QoE risk.

\rev{QoE risk is encoded as a binary variable $y \in \{0,1\}$, with $y=1$ and $y=0$ denoting elevated and non-elevated risk, respectively. We assign each session-level subjective QoE annotation to all temporal windows extracted from the corresponding XR session, formulating QoE-risk estimation as a weakly supervised temporal learning problem over packet-level traffic observations.} The task minimizes
\begin{equation}
\label{eq:qoe_loss}
\mathcal{L}_{\text{qoe}}
=
\mathbb{E}\!\left[
\mathcal{D}\!\left(y,\; f_{\text{qoe}}(\mathbf{X})\right)
\right],
\end{equation}
where $\mathcal{D}(\cdot,\cdot)$ is binary cross-entropy.

The XR predictive modeling problem is formulated as learning two functions,
$f_{\text{traf}}^{*}$ and $f_{\text{qoe}}^{*}$, which minimize their respective losses:
$$
f_{\text{traf}}^{*} = \arg\min_{f_{\text{traf}}} \mathcal{L}_{\text{traf}}, \qquad
f_{\text{qoe}}^{*} = \arg\min_{f_{\text{qoe}}} \mathcal{L}_{\text{qoe}}.
$$

\section{Residual Learning Framework}
\label{sec:reslearn}

\begin{figure}
    \centering
    \includegraphics[trim={0 2 2 0},width=0.8\linewidth]{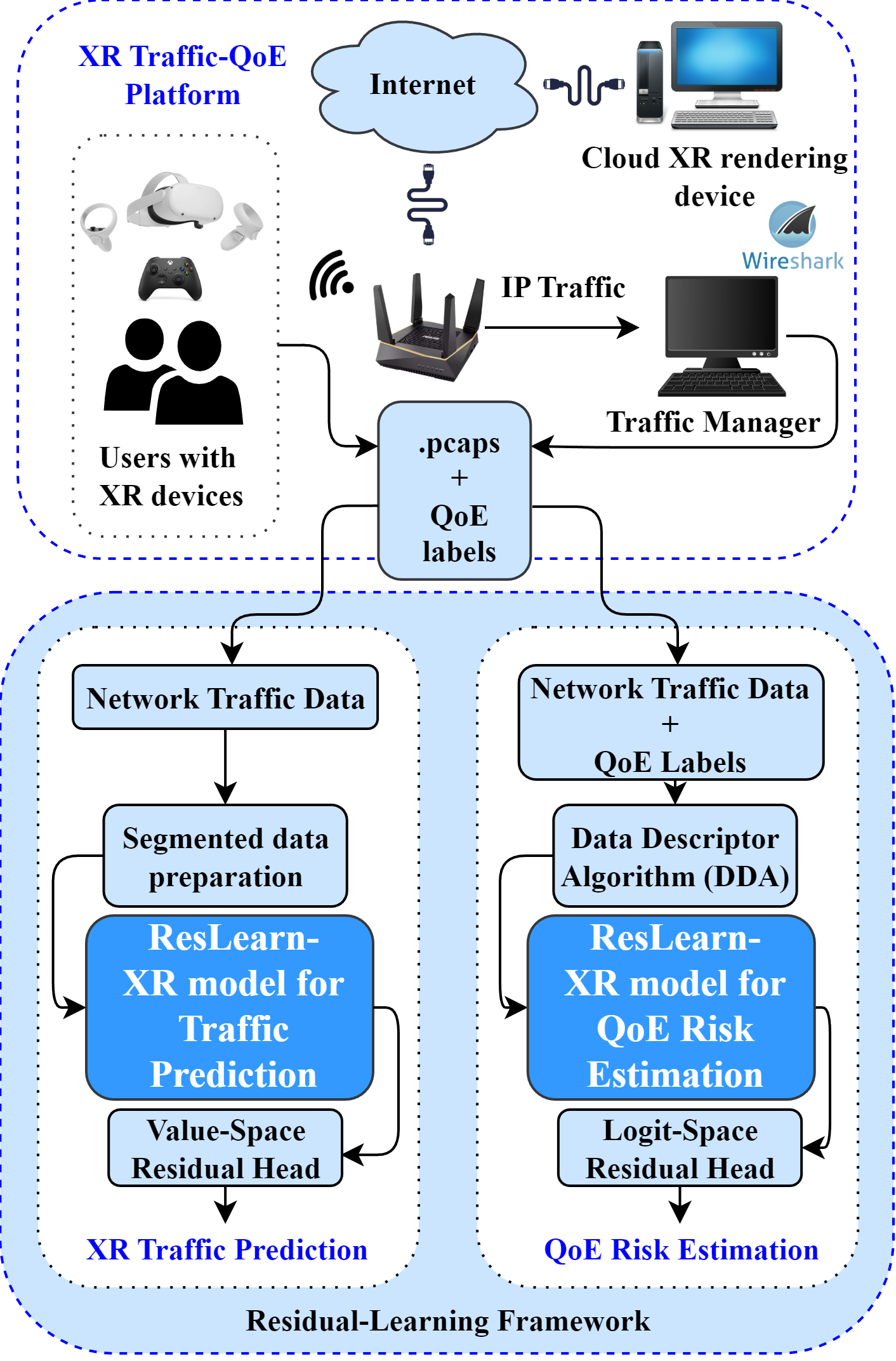}
    \caption{System model of ResLearn-XR framework and XR Traffic-QoE platform.}
    \label{fig:sm}
\end{figure}

As illustrated in Fig.~\ref{fig:sm}, the ResLearn-XR platform interconnects XR head-mounted displays, wireless access networks, and cloud-rendering servers. Bidirectional IP traffic is passively captured per session as packet captures (pcap) traces, and user-reported session-level QoE labels (comfort, tolerable, discomfort, cybersickness) are associated with the corresponding traffic traces. On the traffic prediction path, frame count, average frame size, and inter-arrival interval are derived from packet-level observations, capturing delivery rate, buffering delay, and jitter, respectively, yielding the $m=3$ targets for~\eqref{eq:traffic_prediction_mapping}. On the QoE risk estimation path, the same traces are processed by the \rev{DDA}, which derives frame-timing-aware descriptors from application-layer observables (packet length, direction, inter-arrival time, timestamp) aligned with the session's configured refresh rate, enabling both tasks from a single shared traffic source.

\subsection{XR Traffic Prediction}
\label{subsec:traffic}


ResLearn-XR performs XR traffic prediction with a two-stage model that forecasts frame-aligned traffic metrics derived from packet-level observations and relevant to the MTP latency constraint. These metrics aggregate packet measurements over temporal intervals synchronized with the XR frame-generation period, so that each feature vector corresponds to one rendered-frame interval. The selected metrics---average frame size, frame count, and average frame inter-arrival interval---capture traffic volume, generation rate, and timing, respectively, which are the traffic dimensions most directly related to end-to-end latency.

The segmented data preparation module in Fig.~\ref{fig:sm} uses the Frame Identification Algorithm (FIA)~\cite{11087572} to aggregate packet-level network data over short, non-overlapping temporal windows:
\begin{equation}
\mathbf{z}_k = \mathrm{FIA}\!\left(\{\mathbf{x}_{t} \mid t \in \mathcal{W}_k\}\right),
\end{equation}
where $\mathcal{W}_k = [(k-1)W, kW)$ denotes the $k$-th window of duration $W$, and $\mathrm{FIA}(\cdot)$ is the frame identification and aggregation operator. Applying FIA over an XR session yields
\begin{equation}
\mathbf{Z} = \{\mathbf{z}_k \in \mathbb{R}^{3} \mid k = 1, 2, \ldots, K\},
\end{equation}
where each vector $\mathbf{z}_k = [\, s_k,\; c_k,\; \iota_k \,]$ contains the average frame size $s_k$, frame count $c_k$, and average frame inter-arrival interval $\iota_k$ for the corresponding window.

Let $c_k$ denote the number of XR frames identified by FIA in window $\mathcal{W}_k$. For each frame $n \in \{1,\ldots,c_k\}$, FIA estimates the frame size $S_{k,n}$ from the aggregate payload size and the delivery timestamp $\tau_{k,n}$ from packet-level timestamps. The average frame size and average inter-arrival interval are then
\begin{equation}
\label{eq:avg_frame_size}
s_k =
\begin{cases}
\frac{1}{c_k}\sum_{n=1}^{c_k} S_{k,n}, & c_k > 0,\\[4pt]
0, & c_k = 0,
\end{cases}
\end{equation}
\begin{equation}
\label{eq:avg_frame_iat}
\iota_k =
\begin{cases}
\frac{1}{c_k - 1}\sum_{n=2}^{c_k} \big(\tau_{k,n} - \tau_{k,n-1}\big), & c_k > 1,\\[4pt]
0, & c_k \le 1.
\end{cases}
\end{equation}

The traffic prediction function in~\eqref{eq:traffic_prediction_mapping} is instantiated as a two-stage residual model,
\begin{equation}
\label{eq:ftraf_final}
f_{\text{traf}}
=
f_{\theta}^{(\mathrm{traf})}
+
g_{\phi}^{(\mathrm{traf})},
\end{equation}
where $f_{\theta}^{(\mathrm{traf})}$ denotes the base traffic prediction function and $g_{\phi}^{(\mathrm{traf})}$ denotes the residual correction function. Given a look-back length $S$, the rolling input sequence is
\begin{equation}
\mathbf{Z}^{(S)}_k = [\,\mathbf{z}_{k-S+1}, \ldots, \mathbf{z}_k\,] \in \mathbb{R}^{S \times 3},
\end{equation}
where $\mathbf{z}_k$ denotes the frame-aligned traffic vector for the $k$-th window, and the prediction target is the next-step traffic state $\mathbf{y}_{k+1}=\mathbf{z}_{k+1}$. The base predictor is
\begin{equation}
\widehat{\mathbf{y}}^{(\mathrm{base})}_{k+1}
=
f_{\theta}^{(\mathrm{traf})}\!\left(\mathbf{Z}^{(S)}_k\right),
\end{equation}
where $f_{\theta}^{(\mathrm{traf})}$ is parameterized by $\theta$. ResLearn-XR uses a Transformer encoder as the default base model to capture long-range dependencies through self-attention, but the residual formulation is model-agnostic and can also be instantiated with LSTM, GRU, or Stacked LSTM encoders for controlled comparison.

The base predictor is trained with the mean-squared error (MSE) objective
\begin{equation}
\mathcal{L}_{\mathrm{traf}}^{(\mathrm{base})} =
\mathbb{E}\!\left[
\left\|
\mathbf{y}_{k+1} - \widehat{\mathbf{y}}^{(\mathrm{base})}_{k+1}
\right\|_2^2
\right],
\end{equation}
where $\widehat{\mathbf{y}}^{(\mathrm{base})}_{k+1}$ is the base prediction, $\mathbf{y}_{k+1}$ is the FIA-derived ground-truth traffic vector, and the expectation is taken over the empirical training distribution.

Although the MSE objective provides stable regression, it biases the base predictor toward conditional-mean behavior and attenuates short-term burst dynamics. Bursts are characterized by increased average frame size $s_k$ and frame count $c_k$, often with reduced inter-arrival interval $\iota_k$; underestimating these patterns weakens proactive congestion management and increases MTP-latency risk. As motivated in Appendix~\ref{appendix:mtp_relation}, ResLearn-XR therefore introduces a value-space residual learner $g_{\phi}^{(\mathrm{traf})}$ to model the remaining prediction error in the second stage of~\eqref{eq:traffic_prediction_mapping}.

\rev{With look-back length $S$, the first base prediction is generated at $k=S$:}
\[
{\color{black}
\widehat{\mathbf{y}}^{(\mathrm{base})}_{S+1}
=
f_{\theta}^{(\mathrm{traf})}(\mathbf{Z}^{(S)}_S).
}
\]
\rev{After observing $\mathbf{y}_{S+1}$, the first residual becomes available:}
\[
{\color{black}
\mathbf{e}_{S+1}
=
\mathbf{y}_{S+1}
-
\widehat{\mathbf{y}}^{(\mathrm{base})}_{S+1}.
}
\]
\rev{Consequently, residual refinement begins with the prediction of $\mathbf{y}_{S+2}$, whereas $\mathbf{y}_{S+1}$ is predicted by the base model during residual warm-up. For any subsequent step, the base residual is defined after the corresponding ground-truth traffic state is observed:}
\begin{equation}
\rev{
\mathbf{e}_{k+1}
=
\mathbf{y}_{k+1}
-
\widehat{\mathbf{y}}^{(\mathrm{base})}_{k+1}
\in \mathbb{R}^{m},
}
\end{equation}
\rev{where $m=3$ corresponds to the traffic vector $[s_k,c_k,\iota_k]$. Thus, $\mathbf{e}_{k+1}$ is unavailable when predicting $\mathbf{y}_{k+1}$. To preserve temporal causality, the residual learner uses the most recently observed residual, setting the residual look-back to $S_r=1$; it receives $\tilde{\mathbf{e}}_k$ as input and predicts the next correction $\widehat{\mathbf{r}}_{k+1}$. This one-step residual history avoids undefined early residual entries and maximizes the number of valid residual-training samples for short segmented traffic windows.}

\rev{To form non-negative residual inputs, we compute a global scalar bias from valid training residuals. Let $\mathcal{I}_{\mathrm{tr}}$ denote the training residual indices with $j\ge S+1$. The bias is}
\begin{equation}
\label{eq:residual_bias}
\rev{
b
=
\left|
\min
\left\{
0,
\min_{\substack{j\in\mathcal{I}_{\mathrm{tr}},\, i\in\{1,\ldots,m\}}}
e_{j,i}
\right\}
\right|
\ge 0.
}
\end{equation}
\rev{The scalar $b$ is computed once from the training residuals and remains fixed during validation, testing, and online inference. The bias-shifted residual is}
\begin{equation}
\rev{
\tilde{\mathbf{e}}_{j}
=
\mathbf{e}_{j}+b\mathbf{1}_m,
\quad j=S+1,\ldots,K.
}
\end{equation}
\rev{This shift preserves the temporal ordering of residuals and reduces mixed-sign cancellation when the residual learner models systematic underprediction during burst-dominated periods. Because the supervised target remains $\mathbf{e}_{k+1}$, the predicted correction is expressed in the original residual space before being added to the base prediction.}

\rev{The residual learner estimates the next-step correction as}
\begin{equation}
\rev{
\widehat{\mathbf{r}}_{k+1}
=
g_{\phi}^{(\mathrm{traf})}(\tilde{\mathbf{e}}_k),
\quad k\ge S+1.
}
\end{equation}
\rev{At time $k$, $\tilde{\mathbf{e}}_k$ is observed, whereas $\mathbf{e}_{k+1}$ becomes available only after $\mathbf{y}_{k+1}$ is observed. With the base predictor fixed, the residual learner is trained by minimizing}
\begin{equation}
\rev{
\mathcal{L}^{(\mathrm{res})}_{\mathrm{traf}}
=
\mathbb{E}
\left[
\left\|
\mathbf{e}_{k+1}
-
g_{\phi}^{(\mathrm{traf})}(\tilde{\mathbf{e}}_k)
\right\|_2^2
\right],
\quad k=S+1,\ldots,K-1.
}
\end{equation}

\rev{After residual warm-up, the final next-step traffic prediction is}
\begin{equation}
\rev{
\widehat{\mathbf{y}}_{k+1}
=
\widehat{\mathbf{y}}^{(\mathrm{base})}_{k+1}
+
\widehat{\mathbf{r}}_{k+1},
\quad k\ge S+1.
}
\end{equation}
\rev{For $k=S$, no previously observed residual exists, so the prediction is base-only:}
\[
\rev{
\widehat{\mathbf{y}}_{S+1}
=
\widehat{\mathbf{y}}^{(\mathrm{base})}_{S+1}.
}
\]

\rev{Algorithm~\ref{alg:xrt} summarizes the traffic prediction procedure, where $W$ is the non-overlapping window duration, $S$ is the base-predictor look-back length, and $S_r=1$ is the residual look-back used by the value-space residual learner.} The source code of the XR traffic prediction module is public\footnote{\url{https://github.com/yoga-suhas-km/ResLearn}}.

\begin{algorithm}[t]
\caption{\rev{ResLearn-XR: Two-Stage Traffic Prediction}}
\label{alg:xrt}
\begingroup
\color{black}
\begin{algorithmic}[1]
\Require Packet sequence $\mathbf{X}$, window duration $W$, base look-back $S$, residual look-back $S_r=1$
\Ensure Next-step traffic prediction $\widehat{\mathbf{y}}_{k+1}$

\For{$k=1,\ldots,K$}
    \State $\mathbf{z}_k \gets \mathrm{FIA}(\{\mathbf{x}_t\}_{t\in\mathcal{W}_k})$
    \State $\mathbf{y}_k \gets \mathbf{z}_k$
\EndFor

\State \textbf{Base-stage training:}
\For{$k=S,\ldots,K-1$}
    \State $\mathbf{Z}^{(S)}_k \gets [\mathbf{z}_{k-S+1},\ldots,\mathbf{z}_k]$
    \State $\widehat{\mathbf{y}}^{(\mathrm{base})}_{k+1}
    \gets f_{\theta}^{(\mathrm{traf})}(\mathbf{Z}^{(S)}_k)$
    \State $\mathbf{e}_{k+1} \gets \mathbf{y}_{k+1}
    - \widehat{\mathbf{y}}^{(\mathrm{base})}_{k+1}$
\EndFor

\State Compute $b$ once from valid training residuals:
\[
b=
\left|
\min
\left\{
0,
\min_{\substack{j\in\mathcal{I}_{\mathrm{tr}},\,i\in\{1,\ldots,m\}}}
e_{j,i}
\right\}
\right|.
\]

\For{$j=S+1,\ldots,K$}
    \State $\tilde{\mathbf{e}}_j \gets \mathbf{e}_j + b\mathbf{1}_m$
\EndFor

\State \textbf{Residual-stage training:}
\For{$k=S+1,\ldots,K-1$}
    \State Train $g_{\phi}^{(\mathrm{traf})}$ using $\tilde{\mathbf{e}}_k$ as input and $\mathbf{e}_{k+1}$ as target
\EndFor

\State \textbf{Inference at time $k$:}
\State $\mathbf{Z}^{(S)}_k \gets [\mathbf{z}_{k-S+1},\ldots,\mathbf{z}_k]$
\State $\widehat{\mathbf{y}}^{(\mathrm{base})}_{k+1}
\gets f_{\theta}^{(\mathrm{traf})}(\mathbf{Z}^{(S)}_k)$

\If{$k=S$}
    \State $\widehat{\mathbf{y}}_{k+1} \gets \widehat{\mathbf{y}}^{(\mathrm{base})}_{k+1}$
\Else
    \State $\widehat{\mathbf{r}}_{k+1}
    \gets g_{\phi}^{(\mathrm{traf})}(\tilde{\mathbf{e}}_k)$
    \State $\widehat{\mathbf{y}}_{k+1}
    \gets \widehat{\mathbf{y}}^{(\mathrm{base})}_{k+1}+\widehat{\mathbf{r}}_{k+1}$
\EndIf

\State After $\mathbf{y}_{k+1}$ is observed, compute
$\mathbf{e}_{k+1}=\mathbf{y}_{k+1}-\widehat{\mathbf{y}}^{(\mathrm{base})}_{k+1}$ and update the residual state for the next prediction.
\Return $\widehat{\mathbf{y}}_{k+1}$
\end{algorithmic}
\endgroup
\end{algorithm}

\rev{For a general residual look-back $S_r>1$, residual-stage training begins at $k=S+S_r$, because the earliest available bias-shifted residual is $\tilde{\mathbf{e}}_{S+1}$; with $S_r=1$, training begins at $k=S+1$.}

\subsection{XR QoE Risk Estimation}
\label{subsec:qoe}

\subsubsection{\rev{DDA}}
\label{subsubsec:dda}

The XR QoE risk estimation module estimates cybersickness risk from causal DDA descriptors. \rev{DDA first aggregates packet-level observables (packet size, direction, inter-arrival time, and timestamp) into frame-aligned statistics, and then maps them to temporal descriptors.} These descriptors encode frame-rate deviation from the display target, timing and throughput instability, and offered load as a congestion-delay proxy for supervised QoE risk estimation.


DDA produces one descriptor per non-overlapping temporal window
$\mathcal{W}_k=[(k-1)W,\,kW)$, $k=1,\ldots,T$. Each descriptor is causal, using
only packet observations with timestamps up to $kW$; rolling statistics use a
fixed look-back of $U$ descriptor indices, independent of $W$.


For window $k$, FIA identifies $c_k$ frames with sizes
$\{S_{k,i}\}_{i=1}^{c_k}$ and inter-frame intervals
$\{\Delta\tau_{k,i}\}_{i=2}^{c_k}$, as in
\eqref{eq:avg_frame_size}--\eqref{eq:avg_frame_iat}; $s_k$ and $\iota_k$ follow
\eqref{eq:avg_frame_size} and \eqref{eq:avg_frame_iat}. These FIA-derived
quantities define descriptors of frame delivery, timing instability, and load
conditions relevant to QoE risk.


The effective frame delivery rate is
\begin{equation}
f_k = \frac{c_k}{W},
\end{equation}
where deviations from the display target indicate under-delivery and irregular
frame pacing.


The offered load is
\begin{equation}
r_k = \frac{8}{W \cdot 10^6} \sum_{i=1}^{c_k} S_{k,i},
\end{equation}
where $S_{k,i}$ is the size (bytes) of the $i$-th frame; $r_k$ (Mbps) proxies
congestion pressure and queueing-delay variability.



Inter-frame timing dispersion is
\begin{equation}
j_k =
\begin{cases}
\sqrt{\frac{1}{c_k - 1} \sum_{i=2}^{c_k}
\big(\Delta\tau_{k,i} - \iota_k\big)^2}, & c_k > 1, \\[4pt]
0, & c_k \le 1.
\end{cases}
\end{equation}
Together, $\{f_k,\, s_k,\, \iota_k,\, r_k,\, j_k\}$ causally characterize
delivery rate, payload intensity, pacing regularity, offered load, and timing
variability, whose deviations increase MTP-latency and QoE-degradation risk.


Let $\mathcal{R}$ denote the candidate refresh-rate set; in XR sessions,
$\mathcal{R}=\{60,90,120\}$~Hz. Because the display refresh target $\hat r$ is
not directly observable from network traces, DDA estimates it as
\begin{equation}
\label{eq:qoe_rhat}
\hat r
=
\arg\min_{r\in\mathcal{R}}
\big| r - \mathrm{median}(f_1,\dots,f_T) \big|.
\end{equation}

\rev{For non-standard or adaptive devices, the same estimator can use a device-supported or empirically inferred refresh-rate candidate set.}


Given $\hat r$, pacing descriptors are
\begin{equation}
\rho_k = \frac{f_k}{\hat r}, \qquad
h_k = \frac{\hat r - f_k}{\hat r}, \qquad
d_k = \max\{\hat r - f_k,\,0\},
\end{equation}
where $\rho_k$ measures pacing alignment, $h_k$ relative shortfall, and $d_k$
non-negative under-delivery.


For each scalar sequence $q_k \in \{ r_k, f_k, \iota_k \}$, DDA computes causal
dispersion and trend statistics over the preceding $U$ windows.


\rev{The normalized variability statistic is}
\begin{equation}
\rev{\mathrm{CoV}_{U}(q_k)}
=
\frac{\mathrm{Std}\!\left(q_{k-U+1:k}\right)}
{\mathrm{Mean}\!\left(q_{k-U+1:k}\right) + \varepsilon},
\end{equation}
where $\varepsilon > 0$ ensures numerical stability, and
$\rev{\mathrm{CoV}_{U}(q_k)}$ gives scale-normalized short-term volatility.


Temporal drift over the same horizon is
\begin{equation}
\rev{\beta_{U}(q_k)}
=
\frac{q_k - q_{k-U}}{U}.
\end{equation}
This statistic captures persistent descriptor trends, such as increasing offered
load or degraded frame pacing. Table~\ref{tab:qoe_dda_descriptors} summarizes the
DDA descriptors.



The composite instability indicator is
\begin{equation}
\label{eq:qoe_Jk}
J_k
=
\tanh\!\big(
\lambda_1\,\rev{\mathrm{CoV}_{U}(\iota_k)}
+
\lambda_2\,\rev{\mathrm{CoV}_{U}(f_k)}
+
\lambda_3\,h_k
\big),
\end{equation}
where $\lambda_1$, $\lambda_2$, and $\lambda_3$ are non-negative weights selected
via validation. The terms represent inter-frame timing volatility, frame-rate
volatility, and display-target deviation; $\tanh(\cdot)$ bounds the indicator
while preserving sensitivity to instability.


The DDA output for each session is
\begin{equation}
\mathbf{H} = \{ \mathbf{h}_k \in \mathbb{R}^{12} \mid k = 1,\dots,T \},
\end{equation}
where each $\mathbf{h}_k$ concatenates the twelve features in
Table~\ref{tab:qoe_dda_descriptors}:
\begin{equation}
\begin{split}
\mathbf{h}_k =
\big[
&f_k,\, s_k,\, \iota_k,\, j_k,\, r_k,\, \hat r,\, \rho_k, \\
&h_k,\, d_k,\, \rev{\mathrm{CoV}_{U}(q_k)},\, \rev{\beta_{U}(q_k)},\, J_k
\big],
\end{split}
\end{equation}
where $q_k \in \{ r_k, f_k, \iota_k \}$ denotes the scalar sequence used for the
rolling statistics. For look-back $S$, the causal QoE input is
\begin{equation}
\label{eq:qoe_lookback_sequence}
\mathbf{H}^{(S)}_k
=
[\,\mathbf{h}_{k-S+1}, \dots, \mathbf{h}_k\,]
\in \mathbb{R}^{S \times 12}.
\end{equation}

\begin{table}
\centering
\caption{Causal DDA descriptors used for QoE risk estimation.}
\renewcommand{\arraystretch}{1.03}
\setlength{\tabcolsep}{3pt}
\footnotesize
\begin{adjustbox}{width=\columnwidth}
\begin{tabular}{p{1.25cm} p{2.75cm} p{3.85cm}}
\hline
\textbf{Symbol} & \textbf{Meaning} & \textbf{Role in QoE risk estimation} \\
\hline
\multicolumn{3}{l}{\textbf{A. Base per-window descriptors}} \\
$f_k$   & Effective frame delivery rate
       & Reduced rate increases MTP violation risk \\
$s_k$   & Average payload size per frame
       & Larger frames amplify queueing sensitivity \\
$\iota_k$ & Mean inter-frame interval
         & Irregular spacing degrades motion stability \\
$j_k$   & Inter-frame timing dispersion
       & Elevated values indicate pacing instability \\
$r_k$   & Window-level offered traffic load
       & Higher load increases delay variability risk \\
\hline
\multicolumn{3}{l}{\textbf{B. Pacing relative to display target}} \\
$\hat r$ & Inferred display refresh target
        & Reference timing anchor for pacing alignment \\
$\rho_k$  & Frame pacing ratio ($f_k/\hat r$)
         & Measures alignment with display timing \\
$h_k$     & Normalized pacing shortfall
         & Quantifies relative under-delivery of frames \\
$d_k$     & Non-negative pacing deficit
         & Captures sustained under-pacing events \\
\hline
\multicolumn{3}{l}{\textbf{C. Short-term temporal stability (over $U$ windows)}} \\
$\textcolor{black}{\mathrm{CoV}_{U}(\cdot)}$   & Coefficient of variation
                          & Captures short-term timing and rate volatility \\
$\textcolor{black}{\beta_{U}(\cdot)}$ & Finite-difference trend
                            & Detects persistent temporal drift \\
\hline
\multicolumn{3}{l}{\textbf{D. Composite instability indicator}} \\
$J_k$    & Composite instability index
        & Aggregates correlated timing and pacing cues \\
\hline
\end{tabular}
\end{adjustbox}
\label{tab:qoe_dda_descriptors}
\end{table}

\rev{Section~\ref{subsubsec:dda_ablation} evaluates descriptor-group contributions through an ablation of base per-window, pacing, short-term stability, composite-instability, and full DDA representations.}

\subsubsection{XR QoE Risk Estimation}
\label{subsec:qoe_pred}

Following~\eqref{eq:qoe_mapping}, ResLearn-XR instantiates QoE risk estimation as
a two-stage logit-space model,
\begin{equation}
\label{eq:fqoe_final}
f_{\mathrm{qoe}}
=
\sigma\!\Big(
f_{\theta}^{(\mathrm{qoe})}
+
g_{\phi}^{(\mathrm{qoe})}
\Big),
\end{equation}
where $f_{\theta}^{(\mathrm{qoe})}$ maps DDA-derived descriptors to a base logit,
$g_{\phi}^{(\mathrm{qoe})}$ adds a logit-space residual correction, and
$\sigma(\cdot)$ returns the elevated-risk probability.

\rev{The QoE branch follows the same temporal-causality principle as the traffic-prediction branch: $\mathbf{H}^{(S)}_k=[\mathbf{h}_{k-S+1},\ldots,\mathbf{h}_k]$ contains only observations available up to window $k$, and the residual head uses only the corresponding latent state $\mathbf{s}_k$ at inference. Future descriptors and ground-truth QoE labels are excluded.}

Given $\mathbf{H}^{(S)}_k \in \mathbb{R}^{S \times 12}$ defined in~\rev{\eqref{eq:qoe_lookback_sequence}}, the base-stage estimator produces a scalar logit
\begin{equation}
\ell^{(\mathrm{base})}_k
=
f_{\theta}^{(\mathrm{qoe})}\!\left(\mathbf{H}^{(S)}_k\right)
\in \mathbb{R},
\end{equation}
with base probability
\begin{equation}
\widehat{p}^{(\mathrm{base})}_k
=
\sigma\!\left(\ell^{(\mathrm{base})}_k\right),
\qquad
\sigma(u) = \frac{1}{1+e^{-u}}.
\end{equation}
The base estimator is architecture-agnostic and can be instantiated with LSTM or
GRU encoders. It is trained using class-weighted binary cross-entropy with logits,
\begin{equation}
\label{eq:qoe_base_loss}
\begin{aligned}
\mathcal{L}_{\mathrm{qoe}}^{(\mathrm{base})}
=
\mathbb{E}\!\big[
&-\beta\, y_k \log \sigma\!\left(\ell_k^{(\mathrm{base})}\right) \\
&-(1-y_k)\log\!\left(1-\sigma\!\left(\ell_k^{(\mathrm{base})}\right)\right)
\big],
\end{aligned}
\end{equation}
where $y_k\in\{0,1\}$ is the binary QoE risk label ($1$: elevated risk, $0$:
non-elevated risk), and $\beta=N_0/N_1$ compensates for class imbalance using the
numbers of non-elevated-risk and elevated-risk training samples.



To capture short-duration QoE degradations, let $\mathbf{s}_k$ denote the final-step
latent state produced by the base QoE backbone from $\mathbf{H}^{(S)}_k$. The
residual head maps $\mathbf{s}_k$ to a bounded corrective logit
$\delta_k\in[-n,n]$:
\begin{align}
\ell^{(\mathrm{res})}_k &= \ell^{(\mathrm{base})}_k + \delta_k, \qquad \delta_k = g_{\phi}^{(\mathrm{qoe})}(\mathbf{s}_k), \\
\widehat{p}^{(\mathrm{res})}_k &= \sigma\!\left(\ell^{(\mathrm{res})}_k\right).
\end{align}


During residual-stage training, $f_{\theta}^{(\mathrm{qoe})}$ is frozen and only
$g_{\phi}^{(\mathrm{qoe})}$ is optimized. The residual head is trained directly on
the corrected logit using the same class-weighted loss; no explicit residual
target is required:
\begin{equation}
\label{eq:qoe_res_loss}
\begin{aligned}
\mathcal{L}_{\mathrm{qoe}}^{(\mathrm{res})}
=
\mathbb{E}\!\big[
&-\beta\, y_k \log \sigma\!\left(\ell^{(\mathrm{base})}_k + \delta_k\right) \\
&-(1-y_k)\log\!\left(1-\sigma\!\left(\ell^{(\mathrm{base})}_k + \delta_k\right)\right)
\big],
\end{aligned}
\end{equation}
where $y_k$ and $\beta$ are defined as in~\eqref{eq:qoe_base_loss}. For binary
QoE risk estimation, the predictor outputs
\begin{equation}
\label{eq:fqoe_final_binary}
f_{\mathrm{qoe}}
=
\big[\,\widehat{p}_k,\; 1-\widehat{p}_k\,\big],
\qquad
\widehat{p}_k
=
\sigma\!\left(\ell^{(\mathrm{res})}_k\right),
\end{equation}
where $\widehat{p}_k$ is the uncalibrated elevated-risk probability.


Post-training calibration uses stratum-wise isotonic regression~\cite{pernot2023stratification}
on a held-out split disjoint from base and residual training. Strata depend only
on DDA-derived covariates: the inferred refresh anchor $\hat r$, offered load
$r_k$, pacing ratio $\rho_k$, and composite instability index $J_k$.


Quantile discretizers $\mathcal{B}_r$, $\mathcal{B}_\rho$, and $\mathcal{B}_J$
partition $r_k$, $\rho_k$, and $J_k$ into $K_r$, $K_\rho$, and $K_J$ bins,
inducing the calibration strata
\begin{equation}
\mathcal{K}
=
\Big\{\,
(\hat r,\, b_r,\, b_\rho,\, b_J)
\;:\;\begin{aligned}[t]
&\hat r \in \mathcal{R},\; b_r \in \mathcal{B}_r,\\
&b_\rho \in \mathcal{B}_\rho,\; b_J \in \mathcal{B}_J
\end{aligned}
\,\Big\}.
\end{equation}

Each window $k$ is assigned by
\begin{equation}
\kappa(k)
=
\Phi\!\Big(
\hat r,\;
\mathcal{B}_r(r_k),\;
\mathcal{B}_\rho(\rho_k),\;
\mathcal{B}_J(J_k)
\Big)
\in \mathcal{K},
\end{equation}
which groups samples with comparable load, pacing, and instability conditions.


For stratum $\kappa\in\mathcal{K}$, let $\mathcal{D}_\kappa$ collect calibration
samples with $\kappa(k)=\kappa$. If non-empty, a stratum-specific isotonic map
$\mathcal{C}_\kappa:[0,1]\rightarrow[0,1]$ converts uncalibrated risk
probabilities to calibrated probabilities while preserving monotonicity with
empirical event frequency.


If a fine-grained stratum is empty or unavailable, calibration backs off by
progressively pooling conditioning variables:
\begin{equation}
\label{eq:qoe_calibration_backoff}
\mathcal{C}^{\uparrow}_{\kappa} =
\begin{cases}
\mathcal{C}_{(\hat r,\, b_r,\, b_\rho,\, b_J)}, 
& \mathcal{D}_{(\hat r,\, b_r,\, b_\rho,\, b_J)} \ne \varnothing,\\
\mathcal{C}_{(\hat r,\, b_r,\, b_\rho,\, \star)}, 
& \mathcal{D}_{(\hat r,\, b_r,\, b_\rho,\, \star)} \ne \varnothing,\\
\mathcal{C}_{(\hat r,\, b_r,\, \star,\, \star)}, 
& \mathcal{D}_{(\hat r,\, b_r,\, \star,\, \star)} \ne \varnothing,\\
\mathcal{C}_{(\hat r,\, \star,\, \star,\, \star)}, 
& \mathcal{D}_{(\hat r,\, \star,\, \star,\, \star)} \ne \varnothing,\\
\mathcal{C}_{\mathrm{global}}, 
& \text{otherwise},
\end{cases}
\end{equation}
where ``$\star$'' denotes pooling over that dimension; the most specific available
calibrator is used, with $\mathcal{C}_{\mathrm{global}}$ as the final fallback.

\begin{figure}
 	\centering
 	\includegraphics[trim={0 2 2 0},width=0.9\linewidth]{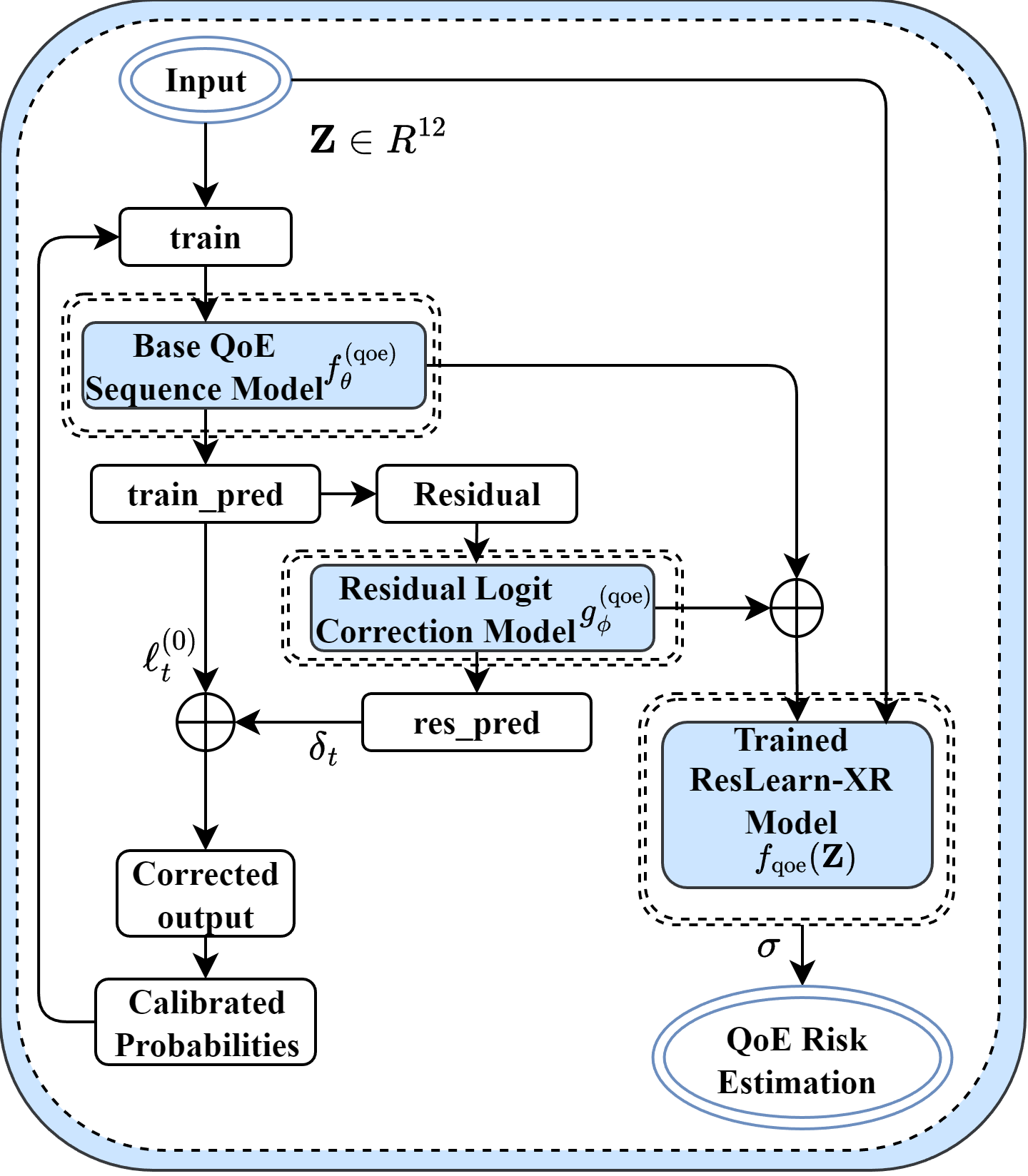} 
 	\caption{Flow of the XR QoE risk estimation module.}
 	\label{fig:qoe_pred}
\end{figure}

At inference, $\tilde{p}_k=\mathcal{C}^{\uparrow}_{\kappa(k)}(\widehat{p}_k)$ and
$f_{\mathrm{qoe}}(\mathbf{H}^{(S)}_k)=[\tilde{p}_k,\;1-\tilde{p}_k]$. The pipeline is shown in Fig.~\ref{fig:qoe_pred} and summarized in
Algorithm~\ref{alg:qoe_full}; the implementation is public\footnote{\url{https://github.com/yoga-suhas-km/XR_QoE_Prediction}}.
\rev{Detailed implementation and reproducibility settings are given in Appendix~\ref{appendix:repro_config}.}


\begin{algorithm}
\caption{ResLearn-XR: QoE Risk Estimation}
\label{alg:qoe_full}
\begin{algorithmic}[1]
\State \textbf{Input:} $\{\mathbf{h}_k\}_{k=1}^{T}$, $S$, $U=16$
\State \textbf{Output:} Calibrated risk vector $[\,\tilde{p}_k,\; 1-\tilde{p}_k\,]$

\For{$k=S,\ldots,T$}
    \State $\mathbf{H}^{(S)}_k \gets [\mathbf{h}_{k-S+1},\ldots,\mathbf{h}_k]$
    \Comment{Causal look-back}
\EndFor

\State \textbf{Base-stage training:}
\For{$k=S,\ldots,T$}
    \State $\ell^{(\mathrm{base})}_k \gets f_{\theta}^{(\mathrm{qoe})}(\mathbf{H}^{(S)}_k)$
    \Comment{Base logit}
\EndFor
\State Train $f_{\theta}^{(\mathrm{qoe})}$ using class-weighted BCE-with-logits
\State Freeze $\theta$

\State \textbf{Residual-stage training:}
\For{$k=S,\ldots,T$}
    \State Obtain final-step latent state $\mathbf{s}_k$ from the frozen base QoE backbone
    \Comment{Causal latent state}
    \State $\delta_k \gets g_{\phi}^{(\mathrm{qoe})}(\mathbf{s}_k)$
    \Comment{Bounded logit correction}
\EndFor
\State Train $g_{\phi}^{(\mathrm{qoe})}$ using class-weighted BCE-with-logits on $\ell^{(\mathrm{base})}_k+\delta_k$

\State \textbf{Calibration:}
\For{$k=S,\ldots,T$}
    \State $\widehat{p}_k \gets \sigma(\ell^{(\mathrm{base})}_k+\delta_k)$
    \Comment{Uncalibrated QoE risk probability}
    \State $\kappa(k) \gets \Phi\!\big(\hat r,\mathcal{B}_r(r_k),\mathcal{B}_\rho(\rho_k),\mathcal{B}_J(J_k)\big)$
    \Comment{Calibration stratum}
\EndFor
\State Fit $\mathcal{C}_\kappa$ with hierarchical back-off to obtain $\mathcal{C}^{\uparrow}_{\kappa}$

\State \textbf{Inference at window $k$:}
\State \hspace{1em} $\widehat{p}_k \gets \sigma(\ell^{(\mathrm{base})}_k+\delta_k)$
\State \hspace{1em} $\tilde{p}_k \gets \mathcal{C}^{\uparrow}_{\kappa(k)}(\widehat{p}_k)$
\State \hspace{1em} \textbf{return} $[\,\tilde{p}_k,\; 1-\tilde{p}_k\,]$
\end{algorithmic}
\end{algorithm}

\section{Experimentation Setup}

\subsection{Experimental Platform and Dataset Collection}
\label{sec:ep_dc}

\begin{figure}
 	\centering
 	\includegraphics[trim={0 2 2 0},width=
    \linewidth]{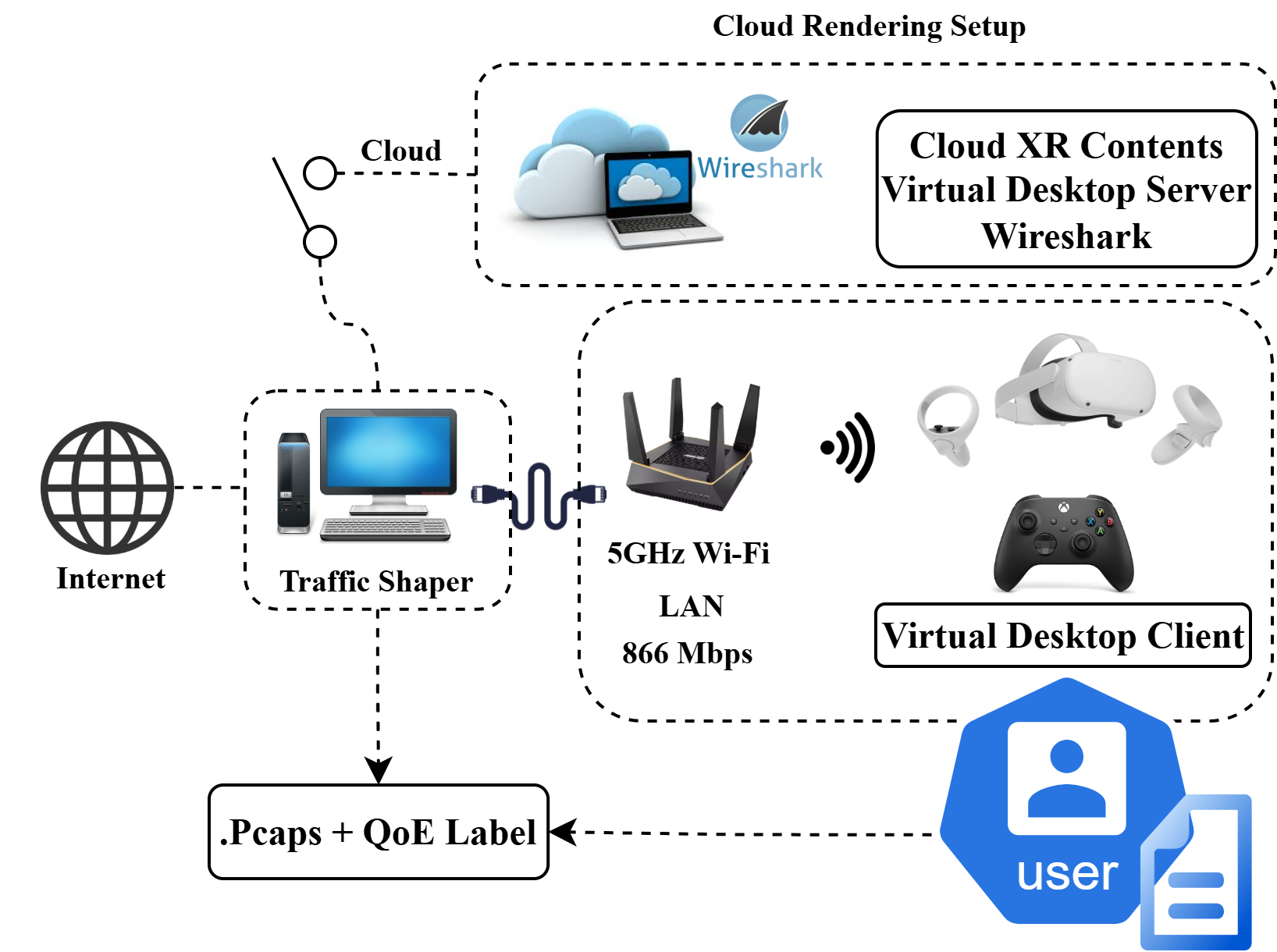} 
 	\caption{Experimental platform for creating XR Traffic-QoE dataset}

 	\label{fig:exp}
\end{figure}

The experimental platform consists of a Meta Oculus Quest~2 head-mounted display (HMD), an Xbox gaming console, a programmable traffic shaping module, and a remote XR content rendering system, as illustrated in Fig.~\ref{fig:exp}. The Quest~2 HMD and Xbox console connect to a local router providing IP connectivity to the Internet, with network access mediated by a traffic shaping module that enforces controlled uplink and downlink rate constraints to emulate bandwidth-limited access network conditions relevant to XR QoE evaluation. The traffic shaping module is implemented on an Ubuntu~18.04 machine using the Wondershaper\footnote{\url{https://github.com/magnific0/wondershaper}}
 and is connected to the university backbone network, which offers an average unconstrained access bandwidth of approximately 120~Mbps.

Remote XR rendering is performed on a cloud-hosted compute instance provided by Paperspace\footnote{\url{https://www.paperspace.com/}}, configured with the Virtual Desktop Streamer (VDS) server application\footnote{\url{https://www.vrdesktop.net/}}
. The corresponding VDS client executes on the Quest~2 HMD and establishes a bidirectional streaming session with the remote rendering instance, such that XR application traffic traverses the Internet between the HMD and the cloud server. Data are captured from a diverse set of commercially available XR services spanning multiple modalities, including VR gaming (Hellblade: Senua’s Sacrifice VR Edition, DiRT Rally~2.0 VR), VR video streaming (Bigscreen VR), social VR with voice communication (VRChat), augmented reality (The Lab—Solar System), and mixed reality (Reality Mixer).



Data collection is conducted with two participants to obtain temporally continuous XR traffic measurements and corresponding session-level QoE labels under controlled network conditions. For each participant, XR sessions are recorded across five bandwidth regimes (15, 30, 60, 120~Mbps, and adaptive), yielding repeated session measurements under identical application and network configurations. 

\rev{After each session, participants provide a four-level ordinal QoE rating aligned with ITU-T P.812/G.1035 and established cybersickness instruments~\cite{kennedy1993ssq, stauffert2023csqvr, kim2018virtual, ITUT_P812, ITU-T2020}: none/comfortable, slight/tolerable, moderate/discomfort, and severe/cybersickness. These session-level ratings are binarized by assigning moderate/discomfort and severe/cybersickness sessions to the elevated-risk class ($y=1$), and none/comfortable and slight/tolerable sessions to the non-elevated-risk class ($y=0$). This protocol reduces participant burden and preserves immersion, but provides weak session-level supervision rather than exact per-window discomfort onset, duration, or intensity. Accordingly, the QoE branch estimates whether traffic windows are associated with elevated session-level QoE risk, not directly measured instantaneous QoE.}

\subsection{Dataset and Experiments}

Table~\ref{table:exp_plan} summarizes the evaluation datasets. XR traffic prediction uses Dataset~I~\cite{data_1} (in-house), Dataset~II~\cite{9685808} (SteamVR traces: SteamVR Home and Beat Saber), and Dataset~III~\cite{questset} (two subsets: diverse VR applications, and a replay under varied network conditions). Dataset~I also supports QoE risk estimation, as no public dataset provides paired continuous traffic traces and session-level QoE labels; its details are in Section~\ref{sec:ep_dc}.

\rev{Traffic prediction follows the effective chronological 40\%/10\%/50\% train/validation/test split within each continuous XR traffic session. Input--target sequences are generated separately within each segment, without shuffling or split-boundary crossing, to prevent look-ahead leakage and evaluate next-step prediction on later unseen traffic.}

\rev{For QoE risk estimation, splitting is performed at the session level because all windows from a session share a weak QoE label. Consistent with Tables~\ref{tab:qoe_u1u2}--\ref{tab:qoe_random}, we report participant-independent evaluation, which trains and tests on complete sessions from different participants, and mixed-user evaluation, which assigns complete sessions from both participants to disjoint training, validation, and testing sets. This design prevents leakage of session-specific patterns and repeated labels; no separate participant-dependent QoE setting is reported.}

\rev{For Dataset~I, all five bandwidth regimes (15, 30, 60, 120~Mbps, and adaptive) are represented in the evaluated splits. Traffic prediction and mixed-user QoE splits are bandwidth-stratified where possible, whereas participant-independent QoE evaluation separates users while retaining matched bandwidth regimes for both participants. This supports evaluation under heterogeneous access-rate conditions and limits dominance by any single bandwidth setting.}

\rev{Informer, FEDformer, and Temporal Fusion Transformer are included as strong traffic-forecasting baselines, while logistic regression, random forest, and gradient-boosted trees serve as non-deep-learning QoE baselines. All models use the corresponding task-specific inputs, splits, and metrics.}

The framework is implemented in Python (NumPy~\cite{harris2020array}, Pandas~\cite{pandas}, Scikit-learn~\cite{sklearn_api}, TensorFlow~\cite{tensorflow2015}, PyTorch~\cite{pytorch}) and trained on an NVIDIA RTX~2080~Super GPU.

\enlargethispage{2\baselineskip}
\begin{table}[!t]
\caption{Datasets and Evaluation Settings for XR Traffic and QoE Risk Prediction}
\centering
\scriptsize
\renewcommand{\arraystretch}{0.9}
\setlength{\tabcolsep}{2pt}
\begin{tabular}{
>{\centering\arraybackslash}m{0.8cm}
>{\centering\arraybackslash}m{0.8cm}
>{\centering\arraybackslash}m{3.2cm}
>{\centering\arraybackslash}m{3.2cm}}
\hline
 & \textbf{Exp.} & \textbf{Services} & \textbf{Applications} \\
\hline
\multicolumn{4}{c}{\textbf{Traffic Prediction}} \\ 
\hline
\rotatebox[origin=c]{90}{\textbf{D-I}} 
& Exp 1 & VR Game, VR Video, VR Chat/VoIP, AR, MR 
& DiRT Rally 2.0, Bigscreen VR, VR Chat, Solar System, Reality Mixer \\ 
\hline
\rotatebox[origin=c]{90}{\textbf{D-II}} 
& Exp 1 & Slow VR Traffic, Fast VR Traffic 
& Steam VR Home, Beat Saber \\ 
\hline
\multirow{2}{*}{\rotatebox[origin=c]{90}{\textbf{III}}} 
& Exp 1 & Fast VR Game 1, Fast VR Game 2, Slow VR Game 1, Slow VR Game 2 
& Beat Saber, Medal of Honor, Forklift Sim, Cooking Sim \\ 
\cline{2-4}
& Exp 2 & Slow VR Traffic, Fast VR Traffic 
& Forklift Sim, Cooking Sim, Beat Saber, Medal of Honor \\
\hline
\hline
\multicolumn{4}{c}{\textbf{QoE Risk Prediction}} \\ 
\hline
\rotatebox[origin=c]{90}{\textbf{D-I}} 
& Exp 1 & VR Game, VR Video, VR Chat/VoIP, AR, MR 
& DiRT Rally 2.0, Bigscreen VR, VR Chat, Solar System, Reality Mixer \\ 
\hline
\end{tabular}
\label{table:exp_plan}
\end{table}

\subsection{Performance Evaluation Metrics}

\subsubsection{Traffic Prediction}
XR traffic prediction performance is evaluated using Root Mean Squared Error (RMSE), Mean Absolute Percentage Error (MAPE), and Symmetric Mean Absolute Percentage Error (SMAPE)~\cite{10437897}. These metrics measure absolute and relative deviations between predicted and observed traffic values.


\subsubsection{QoE Risk Estimation}
QoE risk estimation performance is evaluated using SMAPE, Expected Calibration Error (ECE)~\cite{posocco2021estimating}, and Area Under the Receiver Operating Characteristic Curve (AUC)~\cite{de2022interpreting}.
\rev{For the DDA descriptor ablation, we additionally report Quadratic Weighted Kappa (QWK) and Macro-F1. QWK measures agreement while accounting for the ordinal structure of QoE labels, Macro-F1 evaluates class-balanced prediction performance, and ECE evaluates probability calibration.}

\section{Results and Discussions}

\subsection{Traffic Prediction}
\label{subsec:traffic_results}

\rev{Tables~\ref{tab:p1}--\ref{tab:p4} report the corrected strictly causal traffic-prediction results. Across 16 paired backbone--task comparisons, every ResLearn-XR variant reduces SMAPE relative to its base temporal model, with gains from 0.35\% to 17.84\% and an average gain of approximately 6.7\%. These results support residual correction as a consistent but task-dependent causal refinement of the base forecasters.}

\begin{table}
    \centering
    \begingroup
    \color{black}
    \arrayrulecolor{black}
    \caption{Frame-count prediction performance on Dataset II, comparing baseline temporal models and their ResLearn-XR variants.}
    \scriptsize
    \renewcommand{\arraystretch}{1.2}
    \begin{adjustbox}{width=\columnwidth}
    \begin{tabular}{|l|c|c|c|c|}
    \hline
    \multirow{2}{*}{\textbf{Model}} & \multicolumn{3}{c|}{\textbf{Metrics}} &
    \multirow{2}{*}{\textbf{Causal SMAPE Gain}} \\
    \cline{2-4}
     & \textbf{RMSE} & \textbf{MAPE} & \textbf{SMAPE} & \\
    \hline

    \multicolumn{5}{|c|}{\textbf{Base temporal models}} \\
    \hline
    \textbf{Transformer}  & 0.0427 & 0.0074 & 0.7453 & -- \\
    \textbf{LSTM}         & 0.0420 & 0.0074 & 0.7412 & -- \\
    \textbf{GRU}          & 0.0420 & 0.0074 & 0.7412 & -- \\
    \textbf{Stacked LSTM} & 0.0425 & 0.0075 & 0.7516 & -- \\
    \hline

    \multicolumn{5}{|c|}{\textbf{ResLearn-XR} }\\
    \hline
    \textbf{Transformer}  & \textbf{0.0385} & \textbf{0.0068} & \textbf{0.6821} & \textbf{8.48\%}\\
    \textbf{LSTM}         &  0.0389 & 0.0069 & 0.6931 & 6.49\% \\
    \textbf{GRU}          & 0.0417 & 0.0071 & 0.7162 & 3.37\% \\
    \textbf{Stacked LSTM} & 0.0390 & 0.0069 & 0.6938 & 7.69\% \\
    \hline
    \end{tabular}
    \end{adjustbox}
    \label{tab:p1}
    \arrayrulecolor{black}
    \endgroup
\end{table}


    
    

\begin{table}
    \centering
    \begingroup
    \color{black}
    \arrayrulecolor{black}
    \caption{Frame-size prediction performance on Dataset I, comparing baseline temporal models and their reslearn-xr variants.}
    \small
    \begin{adjustbox}{width=\columnwidth}
    \begin{tabular}{|l|c|c|c|c|}
    \hline
    
    \multirow{2}{*}{\textbf{Model}} & \multicolumn{3}{c|}{\textbf{Metrics}} & 
    \multirow{2}{*}{\textbf{Causal SMAPE Gain}} \\
    \cline{2-4}
    & \textbf{RMSE} & \textbf{MAPE} & \textbf{SMAPE} & \\
    \hline
    
    \multicolumn{5}{|c|}{\textbf{Base temporal models}} \\
    \hline
    \textbf{Transformer}  & 5294.76 & 0.0085 & 0.8506 & -- \\
    \textbf{LSTM}         & 5103.71 & 0.0082 & 0.8195 & -- \\
    \textbf{GRU}          & 5318.15 & 0.0086 & 0.8619 & -- \\
    \textbf{Stacked LSTM} & 5104.09 & 0.0082 & 0.8222 & -- \\
    \hline
    
    \multicolumn{5}{|c|}{\textbf{ResLearn-XR} }\\
    \hline
    \textbf{Transformer}  & \textbf{4760.99} & \textbf{0.0076} & \textbf{0.7600} & \textbf{10.65\%} \\
    \textbf{LSTM}         & 4788.68 & 0.0076 & 0.7663 & 6.49\% \\
    \textbf{GRU}          & 4953.34 & 0.0079 & 0.7968 & 7.54\% \\
    \textbf{Stacked LSTM} & 4987.10 & 0.0080 & 0.8035 & 2.27\% \\
    \hline
    \end{tabular}
    \end{adjustbox}
    \label{tab:p2}
    \arrayrulecolor{black}
    \endgroup
\end{table}

\rev{The gains vary by traffic descriptor and backbone. Table~\ref{tab:p1} shows frame-count SMAPE reductions of 3.37\%--8.48\% on Dataset~II. For frame-size prediction, Tables~\ref{tab:p2} and~\ref{tab:p3} show maximum gains of 10.65\% and 14.21\%, respectively; in Dataset~III Experiment~2, the LSTM residual variant gives the lowest absolute error, whereas the Transformer yields the largest relative gain. Table~\ref{tab:p4} reports the largest overall improvement, 17.84\%, for Transformer-based inter-arrival-time prediction, while the GRU gain is marginal at 0.35\%. Thus, residual learning is best interpreted as a lightweight, task-dependent refinement mechanism.}

\begin{table}
\centering
\begingroup
\color{black}
\arrayrulecolor{black}
\caption{Frame-size prediction performance on Dataset III (Experiment 2), comparing baseline temporal models and their reslearn-xr variants.}
\scriptsize

\renewcommand{\arraystretch}{1.2}
\begin{adjustbox}{width=\columnwidth}
\begin{tabular}{|l|c|c|c|c|}
\hline
\multirow{2}{*}{\textbf{Model}} & \multicolumn{3}{c|}{\textbf{Metrics}} &
\multirow{2}{*}{\textbf{Causal SMAPE Gain}} \\
\cline{2-4}
& \textbf{RMSE} & \textbf{MAPE} & \textbf{SMAPE} & \\
\hline

\multicolumn{5}{|c|}{\textbf{Base temporal models}} \\
\hline
\textbf{Transformer}  & 53573.43 & 0.0051 & 0.5094 & -- \\
\textbf{LSTM}         & 45694.86 & 0.0045 & 0.4445 & -- \\
\textbf{GRU}          & 50025.70 & 0.0046 & 0.4608 & -- \\
\textbf{Stacked LSTM} & 46472.13 & 0.0043 & 0.4258 & -- \\
\hline

\multicolumn{5}{|c|}{\textbf{ResLearn-XR}} \\
\hline
\textbf{Transformer}  & 47578.00 & 0.0044 & 0.4370 & \textbf{14.21\%} \\
\textbf{LSTM}         & \textbf{43514.14} & \textbf{0.0040} & \textbf{0.4036} & 9.20\% \\
\textbf{GRU}          & 48598.58 & 0.0045 & 0.4498 & 2.39\% \\
\textbf{Stacked LSTM} & 44303.85 & 0.0042 & 0.4232 & 0.61\% \\
\hline
\end{tabular}
\end{adjustbox}
\label{tab:p3}
\arrayrulecolor{black}
\endgroup
\end{table}





\begin{table}
    \centering
    \begingroup
    \color{black}
    \arrayrulecolor{black}
    \caption{Frame inter-arrival time prediction performance on Dataset III
    (Experiment 1), comparing baseline temporal models and their reslearn-xr variants.}
    \scriptsize

    \renewcommand{\arraystretch}{1.2}
    \begin{adjustbox}{width=\columnwidth}
    \begin{tabular}{|l|c|c|c|c|}
    \hline
    \multirow{2}{*}{\textbf{Model}} & \multicolumn{3}{c|}{\textbf{Metrics}} &
    \multirow{2}{*}{\textbf{Causal SMAPE Gain}} \\
    \cline{2-4}
    & \textbf{RMSE} & \textbf{MAPE} & \textbf{SMAPE} & \\
    \hline
    
    \multicolumn{5}{|c|}{\textbf{Base temporal models}} \\
    \hline
    \textbf{Transformer}  & 0.1371 & 0.0118 & 1.1752 & -- \\
    \textbf{LSTM}         & 0.1425 & 0.0121 & 1.2032 & -- \\
    \textbf{GRU}          & 0.1309 & 0.0114 & 1.1351 & -- \\
    \textbf{Stacked LSTM} & 0.1393 & 0.0118 & 1.1735 & -- \\
    \hline
    
    \multicolumn{5}{|c|}{\textbf{ResLearn-XR}} \\
    \hline
    \textbf{Transformer}  & \textbf{0.1248} & \textbf{0.0106} & \textbf{0.9655} & \textbf{17.84\%} \\
    \textbf{LSTM}         & 0.1303 & 0.0113 & 1.1264 & 6.38\% \\
    \textbf{GRU}          & 0.1307 & 0.0113 & 1.1311 & 0.35\% \\
    \textbf{Stacked LSTM} & 0.1322 & 0.0113 & 1.1312 & 3.60\% \\
    \hline
    \end{tabular}
    \end{adjustbox}
    \label{tab:p4}
    \arrayrulecolor{black}
    \endgroup
\end{table}


    
    

\begin{table}[t]
\centering
\begingroup
\arrayrulecolor{black}
\caption{\rev{Comparison with Transformer-based forecasting baselines.}}
\scriptsize
\renewcommand{\arraystretch}{1.15}
\begin{adjustbox}{width=\columnwidth}
\begin{tabular}{|l|c|c|c|}
\hline
\rev{\textbf{Model}} &
\rev{\textbf{RMSE}} &
\rev{\textbf{MAPE}} &
\rev{\textbf{SMAPE}} \\
\hline
\rev{Informer} &
\rev{4818.62} &
\rev{0.0093} &
\rev{0.92} \\
\hline
\rev{FEDformer} &
\rev{5676.41} &
\rev{0.0092} &
\rev{0.93} \\
\hline
\rev{Temporal Fusion Transformer} &
\rev{4958.27} &
\rev{0.0085} &
\rev{0.85} \\
\hline
\rev{ResLearn-XR, Transformer backbone} &
\rev{\textbf{4760.99}} &
\rev{\textbf{0.0076}} &
\rev{\textbf{0.76}} \\
\hline
\end{tabular}
\end{adjustbox}
\label{tab:strong_traffic_baselines}
\arrayrulecolor{black}
\endgroup
\end{table}


\rev{Table~\ref{tab:strong_traffic_baselines} compares the Transformer-based ResLearn-XR model with stronger forecasting baselines under the same chronological split. ResLearn-XR attains the lowest RMSE, MAPE, and SMAPE, improving the best competing values from 4818.62 to 4760.99, 0.0085 to 0.0076, and 0.85 to 0.76, respectively. The controlled residual-learning claim remains based on the paired comparisons in Tables~\ref{tab:p1}--\ref{tab:p4}.}

\begin{table}[t]
\centering
\caption{\rev{Approximate MTP-risk detection analysis using a queueing-delay proxy derived from predicted burst traffic.}}
\scriptsize
\renewcommand{\arraystretch}{1.15}
\begin{adjustbox}{width=\columnwidth}
\begin{tabular}{|l|c|c|c|c|}
\hline
\rev{\textbf{Model}} & \rev{\textbf{Precision}} & \rev{\textbf{Recall}} & \rev{\textbf{F1}} & \rev{\textbf{Missed-Risk Rate}} \\
\hline
\rev{Base-only predictor} & \rev{0.78} & \rev{0.63} & \rev{0.70} & \rev{0.37} \\
\hline
\rev{Base + residual predictor} & \rev{0.84} & \rev{0.81} & \rev{0.82} & \rev{0.19} \\
\hline
\end{tabular}
\end{adjustbox}
\label{tab:mtp_risk_mapping}
\end{table}

\rev{Using the queueing proxy in Appendix~\ref{appendix:mtp_relation}, the predicted offered load is estimated as $\widehat{L}_{k}=\widehat{c}_{k}\widehat{s}_{k}/(\widehat{\iota}_{k}+\epsilon)$ and marked as MTP-risk positive when it exceeds a predefined threshold. Table~\ref{tab:mtp_risk_mapping} shows that residual correction increases proxy-risk recall from 0.63 to 0.81 and reduces the missed-risk rate from 0.37 to 0.19. This indicates improved detection of burst-induced risk windows, while direct closed-loop MTP-violation measurement remains future work.}

\rev{Although this study focuses on predictive modeling rather than closed-loop control, ResLearn-XR outputs can support XR-aware management. Traffic predictions provide offered-load and burst-risk indicators for proactive bandwidth, queue, or rate-adaptation decisions, while calibrated QoE-risk probabilities can inform admission control, edge scaling, rendering adaptation, or migration. The bandwidth settings used here define controlled evaluation conditions and are not dynamically adjusted; closed-loop resource orchestration remains future work.}

\begin{table}[t]
\centering
\caption{\rev{Approximate model complexity, FP32 trainable-weight memory, and model-only feasibility comparison.}}
\scriptsize
\renewcommand{\arraystretch}{1.15}
\setlength{\tabcolsep}{1.5pt}
\begin{adjustbox}{width=\columnwidth}
\begin{tabular}{|
>{\raggedright\arraybackslash}p{2.1cm}|
>{\centering\arraybackslash}p{1.5cm}|
>{\centering\arraybackslash}p{1.4cm}|
>{\centering\arraybackslash}p{1.8cm}|
>{\centering\arraybackslash}p{1.2cm}|}
\hline
\rev{\textbf{Model}} & \rev{\textbf{Trainable Parameters}} & \rev{\textbf{FP32 Weight Memory}} & \rev{\textbf{Inference Time / Window}} & \rev{\textbf{Model-only Feasible}} \\
\hline
\rev{Transformer} & \rev{$\sim$265K} & \rev{$\sim$1.06 MB} & \rev{$\sim$0.42 ms} & \rev{Yes} \\
\hline
\rev{Transformer + Residual} & \rev{$\sim$269K} & \rev{$\sim$1.08 MB} & \rev{$\sim$0.47 ms} & \rev{Yes} \\
\hline
\rev{LSTM} & \rev{$\sim$68K} & \rev{$\sim$0.27 MB} & \rev{$\sim$0.31 ms} & \rev{Yes} \\
\hline
\rev{LSTM + Residual} & \rev{$\sim$72K} & \rev{$\sim$0.29 MB} & \rev{$\sim$0.36 ms} & \rev{Yes} \\
\hline
\rev{GRU} & \rev{$\sim$51K} & \rev{$\sim$0.20 MB} & \rev{$\sim$0.28 ms} & \rev{Yes} \\
\hline
\rev{GRU + Residual} & \rev{$\sim$55K} & \rev{$\sim$0.22 MB} & \rev{$\sim$0.33 ms} & \rev{Yes} \\
\hline
\rev{Stacked LSTM} & \rev{$\sim$200K} & \rev{$\sim$0.80 MB} & \rev{$\sim$0.55 ms} & \rev{Yes} \\
\hline
\rev{Stacked LSTM + Residual} & \rev{$\sim$204K} & \rev{$\sim$0.82 MB} & \rev{$\sim$0.61 ms} & \rev{Yes} \\
\hline
\end{tabular}
\end{adjustbox}
\vspace{1.1em}
\begin{minipage}{\columnwidth}
\scriptsize\emph{\rev{Note:}} \rev{Feasibility refers to model-only inference and excludes packet capture, descriptor extraction, controller communication, scheduling, and actuation delay.}
\end{minipage}
\label{tab:complexity}
\end{table}

\rev{Table~\ref{tab:complexity} reports approximate trainable parameters, FP32 weight memory, and per-window inference latency. The residual head adds only about 4K parameters and 0.02~MB to each backbone; for the Transformer, latency increases from 0.42~ms to 0.47~ms, and all recurrent variants remain below 1~ms. These values support low-latency model-only inference, but they do not include packet capture, descriptor extraction, controller communication, scheduling, or actuation delay; therefore, full deployment still requires system-level profiling and closed-loop validation.}

\subsection{QoE Risk Estimation}
\label{subsubsec:qoe_perf}

\begin{figure}[t]
  \centering
  \subfloat[DiRTRally 2.0 (15 Mbps, 60 Hz)\label{fig:traj_dirtrally}]
  {\includegraphics[width=0.49\linewidth]{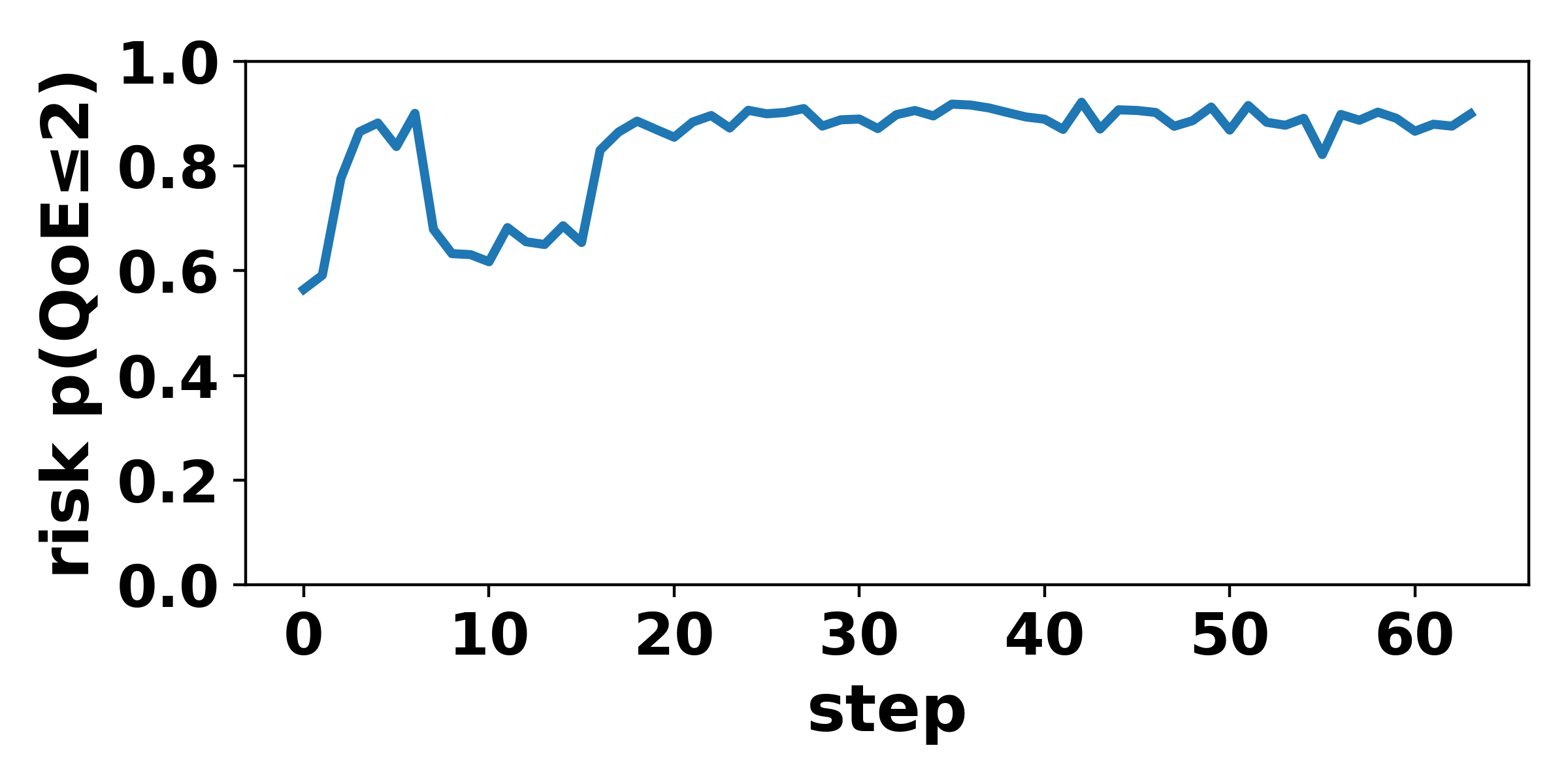}}\hfill
  \subfloat[RealityMixer (30 Mbps, 120 Hz)\label{fig:traj_realitymixer}]
  {\includegraphics[width=0.49\linewidth]{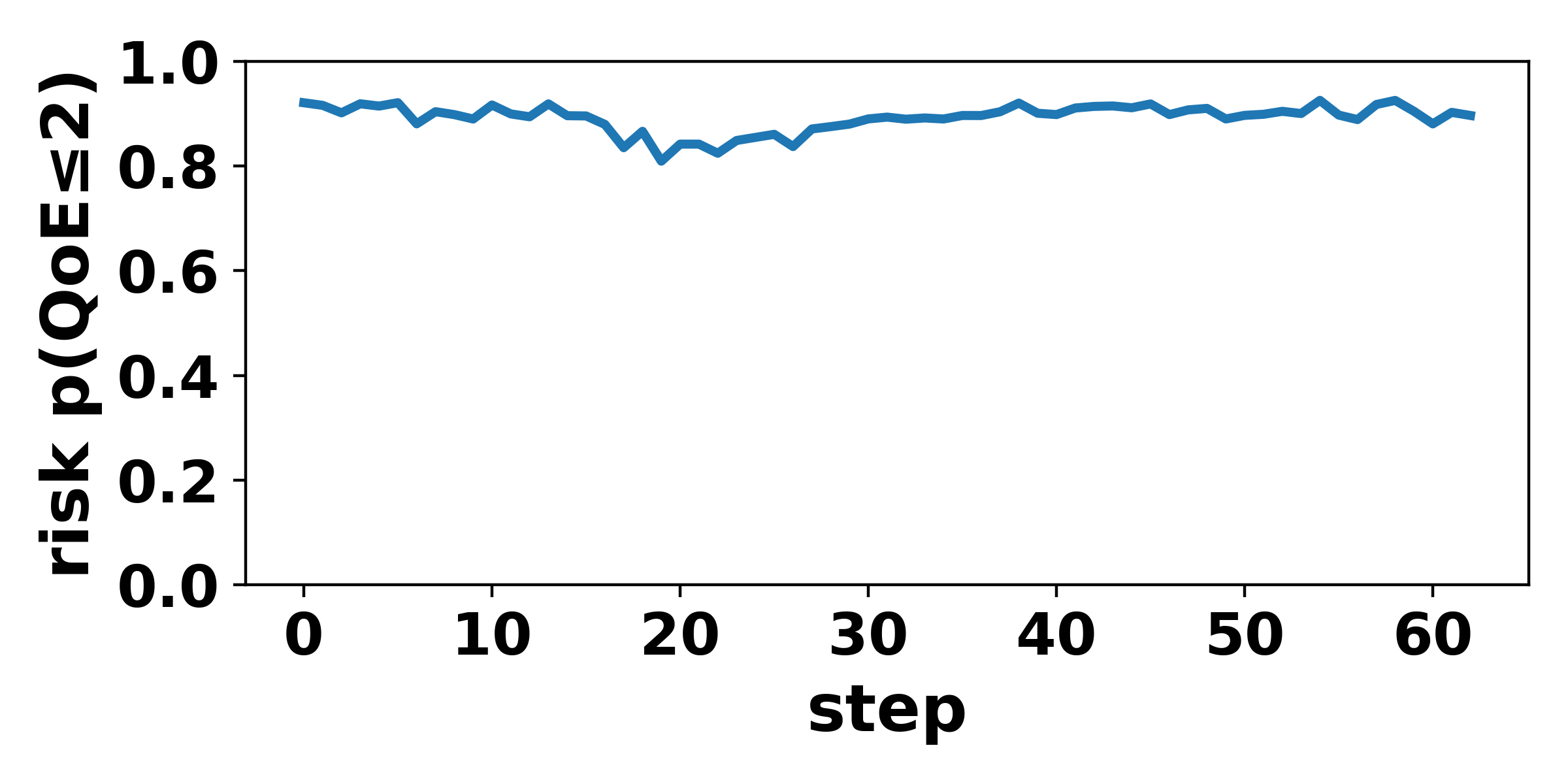}}\hfill
  \subfloat[The Lab (AB cloud, 60 Hz)\label{fig:traj_thelab}]
  {\includegraphics[width=0.49\linewidth]{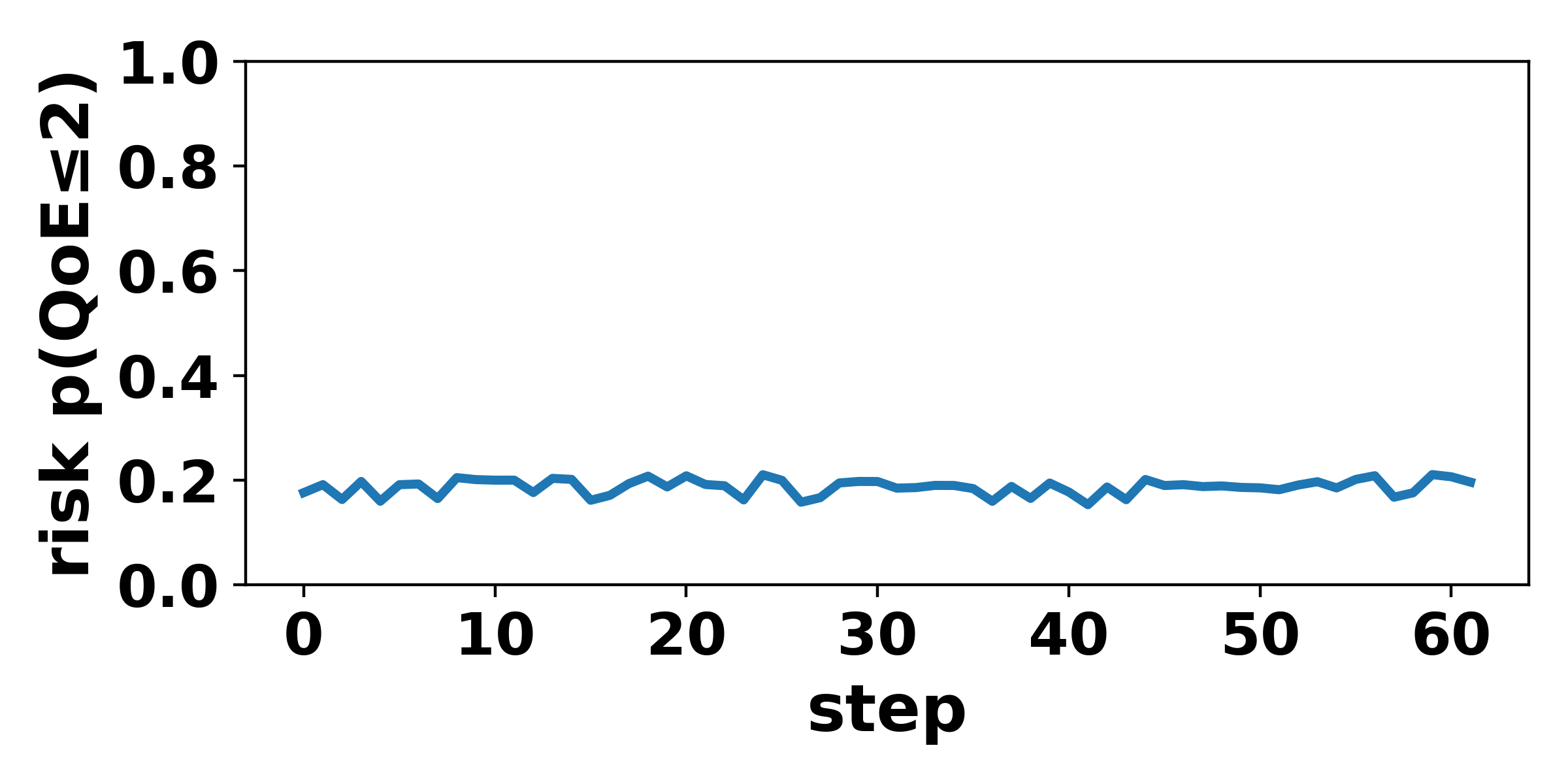}}\hfill
  \subfloat[Bigscreen (120 Mbps, 90 Hz)\label{fig:traj_bigscreen}]
  {\includegraphics[width=0.49\linewidth]{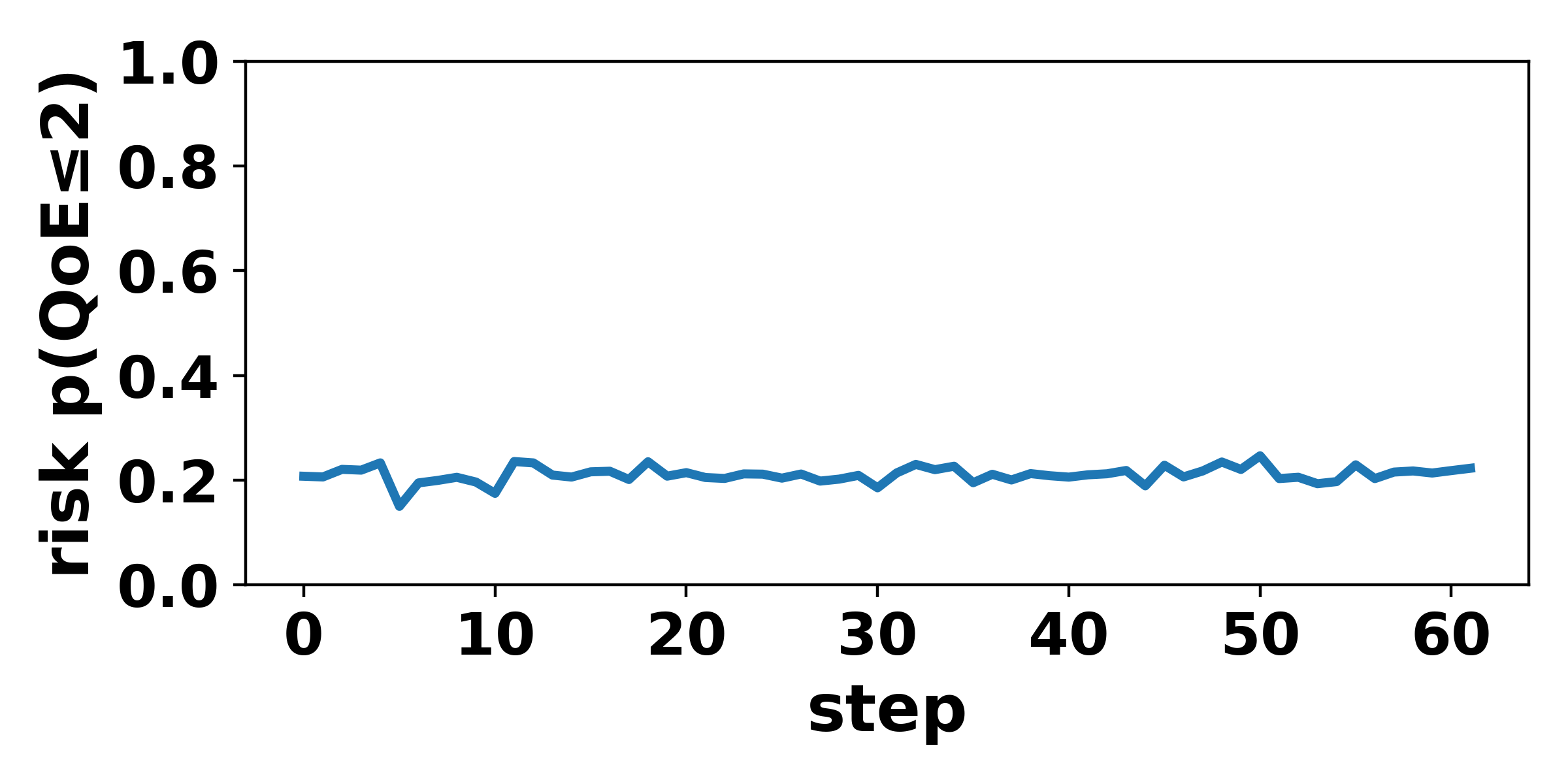}}\hfill
  \subfloat[Hellblade (30 Mbps, 120 Hz)\label{fig:traj_hellblade}]
  {\includegraphics[width=0.49\linewidth]{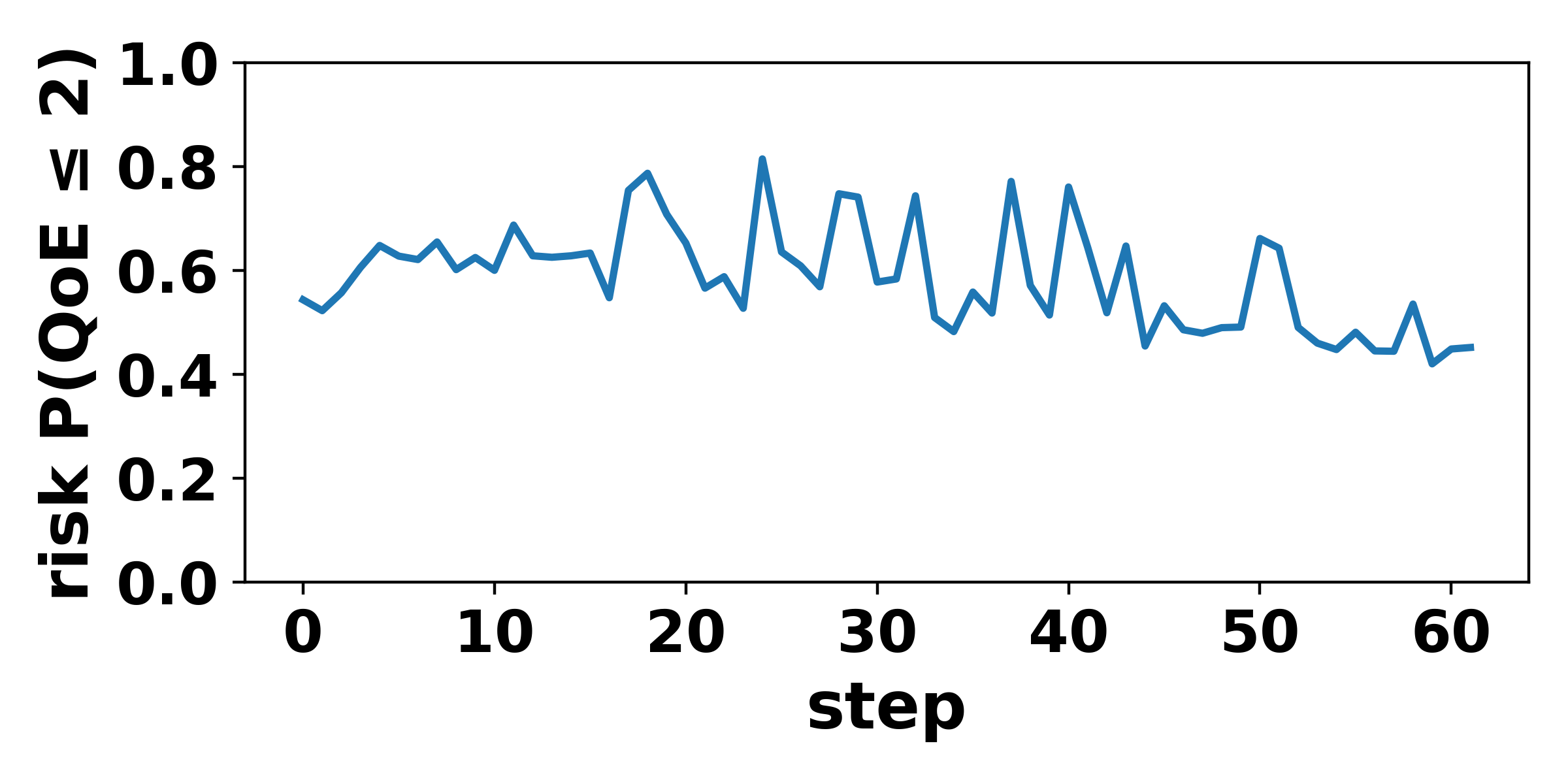}}\hfill
  \subfloat[The Lab (60 Mbps, 90 Hz)\label{fig:traj_good}]
  {\includegraphics[width=0.49\linewidth]{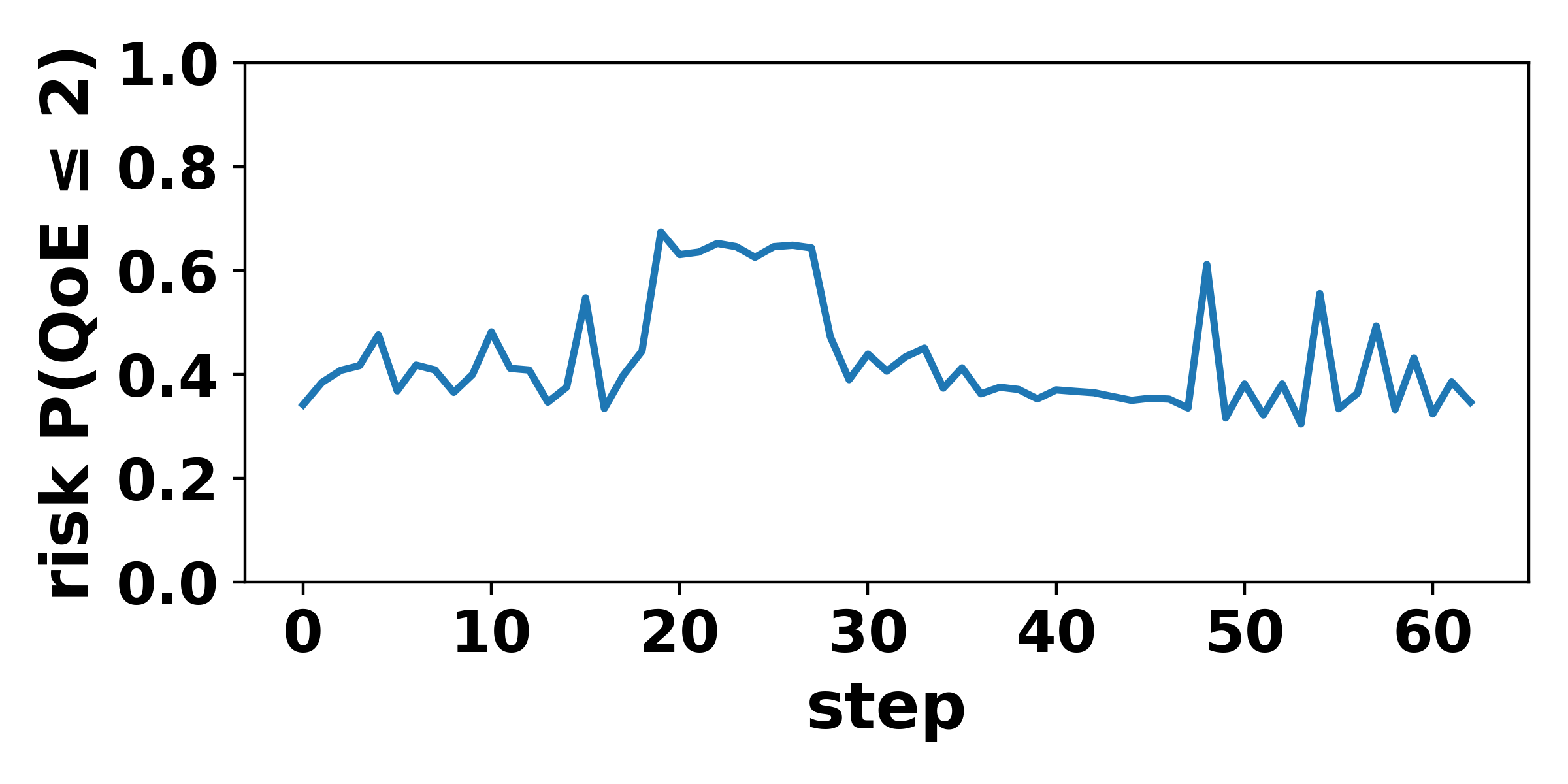}}

  \caption{Predicted QoE risk trajectories $p(\mathrm{QoE}\le2)$ for representative XR sessions under varying applications and network conditions. (a), (b), and (e) correspond to sessions associated with cybersickness, while (c), (d), and (f) correspond to low-risk sessions. Trajectories in (a)–(d) are produced by ResLearn-XR, whereas (e) and (f) are produced by non-ResLearn-XR baseline models.}

  \label{fig:qoe_risk_traj}
\end{figure}

QoE risk estimation evaluates probabilistic prediction of the elevated-risk event $\mathrm{QoE}\le2$ under user and network variability. Tables~\ref{tab:qoe_u1u2}--\ref{tab:qoe_random} and Fig.~\ref{fig:qoe_risk_traj} summarize prediction error, calibration, discrimination, and temporal risk behavior.

Under participant-independent evaluation, ResLearn-XR reduces SMAPE for all reported backbones. The Transformer residual model decreases SMAPE from 1.250 to 0.259 for U2$\rightarrow$U1 and from 1.245 to 0.152 for U1$\rightarrow$U2, while increasing AUC to 0.874 and 0.930, respectively. In the mixed-user setting, residual models also reduce SMAPE; the LSTM variant gives the lowest error (0.429), while the Transformer gives the highest AUC (0.909).

\rev{Calibration is split- and backbone-dependent. For example, in Table~\ref{tab:qoe_u2u1}, the non-residual Transformer and Stacked LSTM have lower ECE (0.052) than the residual Transformer (0.076), whereas the residual Transformer improves ECE in Table~\ref{tab:qoe_u1u2}. Therefore, SMAPE, AUC, and ECE are interpreted as complementary rather than uniformly aligned model properties.}

Figure~\ref{fig:qoe_risk_traj} qualitatively supports these trends. ResLearn-XR produces sustained high-risk trajectories for sessions associated with cybersickness and low-variance trajectories for comfortable sessions, whereas the displayed non-residual baselines are noisier and less separable.

\rev{Because the QoE dataset contains two participants and session-level weak labels, the participant-independent results should be interpreted as preliminary cross-subject evidence rather than broad user-level generalization. The results support the feasibility of traffic-based QoE risk estimation under weak supervision, while larger cohorts, temporally localized feedback, and broader XR scenarios are required for robust user-independent validation.}

\begin{table}[t]
\centering
\caption{\rev{Comparison with non-deep-learning QoE risk baselines under
participant-independent evaluation.}}
\scriptsize
\renewcommand{\arraystretch}{1.15}
\begin{tabular}{|l|c|c|c|}
\hline
\rev{\textbf{Model}} &
\rev{\textbf{AUC}} &
\rev{\textbf{ECE}} &
\rev{\textbf{Macro-F1}} \\
\hline
\rev{Logistic regression} &
\rev{0.681} &
\rev{0.146} &
\rev{0.382} \\
\hline
\rev{Random forest} &
\rev{0.754} &
\rev{0.113} &
\rev{0.421} \\
\hline
\rev{Gradient-boosted trees} &
\rev{0.812} &
\rev{0.089} &
\rev{0.458} \\
\hline
\rev{\textbf{ResLearn-XR}} &
\rev{\textbf{0.930}} &
\rev{\textbf{0.057}} &
\rev{\textbf{0.497}} \\
\hline
\end{tabular}
\label{tab:classical_qoe_baselines}
\end{table}

\rev{Table~\ref{tab:classical_qoe_baselines} shows that ResLearn-XR attains the highest AUC and Macro-F1 and the lowest ECE among the non-deep-learning baselines, indicating that temporal residual modeling adds value beyond the DDA-derived descriptors alone.}

\begin{table}
\centering
\caption{QoE risk estimation performance under participant-independent evaluation (U2 train / U1 test).}
\scriptsize

\renewcommand{\arraystretch}{1.2}
\begin{adjustbox}{width=\columnwidth}
\begin{tabular}{|l|c|c|c|c|}
\hline
\multirow{2}{*}{} & \multicolumn{3}{c|}{\textbf{Metrics}} & \\
\cline{2-4}
\textbf{Model} & \textbf{SMAPE} & \textbf{ECE} & \textbf{AUC} & \textbf{\% SMAPE Improvement} \\
\hline
\multicolumn{5}{|c|}{\textbf{Non-ResLearn-XR Algorithm}} \\
\hline
\textbf{Transformer}  & 1.250 & 0.283 & 0.764 &  \\
\textbf{LSTM}         & 1.259 & 0.170 & 0.679 &  \\
\textbf{GRU}          & 1.114 & 0.462 & 0.571 &  \\
\textbf{Stacked LSTM} & 1.167 & 0.269 & 0.752 &  \\
\hline
\multicolumn{5}{|c|}{\textbf{ResLearn-XR Solution}} \\
\hline
\textbf{Transformer}  & \textbf{0.259} & \textbf{0.119} & \textbf{0.874} & \textbf{79.28\%} \\
\textbf{LSTM}         & 0.662 & 0.329 & 0.679 & 47.42\% \\
\textbf{GRU}          & 0.923 & 0.267 & 0.571 & 17.15\% \\
\textbf{Stacked LSTM} & 0.528 & 0.247 & 0.752 & 54.76\% \\
\hline
\end{tabular}
\end{adjustbox}
\label{tab:qoe_u1u2}
\end{table}

\begin{table}[t]
\centering
\caption{QoE risk estimation performance under participant-independent evaluation (U1 train / U2 test).}
\scriptsize

\renewcommand{\arraystretch}{1.2}
\begin{adjustbox}{width=\columnwidth}
\begin{tabular}{|l|c|c|c|c|}
\hline
\multirow{2}{*}{} & \multicolumn{3}{c|}{\textbf{Metrics}} & \\
\cline{2-4}
\textbf{Model} & \textbf{SMAPE} & \textbf{ECE} & \textbf{AUC} & \textbf{\% SMAPE Improvement} \\
\hline
\multicolumn{5}{|c|}{\textbf{Non-ResLearn-XR Algorithm}} \\
\hline
\textbf{Transformer}  & 1.245 & \textbf{0.052} & 0.500 &  \\
\textbf{LSTM}         & 1.068 & 0.084 & 0.500 &  \\
\textbf{GRU}          & 1.347 & 0.496 & 0.390 &  \\
\textbf{Stacked LSTM} & 1.245 & \textbf{0.052} & 0.500 &  \\
\hline
\multicolumn{5}{|c|}{\textbf{ResLearn-XR Solution}} \\
\hline
\textbf{Transformer}  & \textbf{0.152} & 0.076 & \textbf{0.930} & \textbf{87.79\%} \\
\textbf{LSTM}         & 0.507 & 0.253 & 0.706 & 52.53\% \\
\textbf{GRU}          & 0.369 & 0.154 & 0.821 & 72.61\% \\
\textbf{Stacked LSTM} & 0.709 & 0.354 & 0.539 & 43.05\% \\
\hline
\end{tabular}
\end{adjustbox}
\label{tab:qoe_u2u1}
\end{table}

\begin{table}
\centering
\caption{QoE risk estimation performance under mixed-user evaluation.}
\scriptsize

\renewcommand{\arraystretch}{1.2}
\begin{adjustbox}{width=\columnwidth}
\begin{tabular}{|l|c|c|c|c|}
\hline
\multirow{2}{*}{} & \multicolumn{3}{c|}{\textbf{Metrics}} & \\
\cline{2-4}
\textbf{Model} & \textbf{SMAPE} & \textbf{ECE} & \textbf{AUC} & \textbf{\% SMAPE Improvement} \\
\hline
\multicolumn{5}{|c|}{\textbf{Non-ResLearn-XR Algorithm}} \\
\hline
\textbf{Transformer}  & 1.279 & 0.296 & 0.563 &  \\
\textbf{LSTM}         & 1.211 & 0.365 & 0.656 &  \\
\textbf{GRU}          & 0.603 & 0.258 & 0.760 &  \\
\textbf{Stacked LSTM} & 1.216 & 0.241 & 0.612 &  \\
\hline
\multicolumn{5}{|c|}{\textbf{ResLearn-XR Solution}} \\
\hline
\textbf{Transformer}  & 0.817 & 0.179 & \textbf{0.909} & 36.12\% \\
\textbf{LSTM}         & \textbf{0.429} & 0.214 & 0.802 & \textbf{64.57\%} \\
\textbf{GRU}          & 0.500 & \textbf{0.168} & 0.875 & 17.08\% \\
\textbf{Stacked LSTM} & 1.167 & 0.357 & 0.802 & 4.03\% \\
\hline
\end{tabular}
\end{adjustbox}

\label{tab:qoe_random}
\end{table}

\subsection{Effect of the DDA Descriptors}
\label{subsubsec:dda_ablation}

To evaluate the contribution of the proposed DDA, we compare QoE risk estimation performance with base per-window features alone and with the full DDA descriptor set. Results are reported in Table~\ref{tab:dda_effect} for the participant-independent configuration trained on User~1 and evaluated on User~2.

Incorporating DDA descriptors yields consistent improvements across all evaluation metrics. Relative to the base per-window features, SMAPE decreases from 0.2831 to 0.1524, ECE decreases from 0.1399 to 0.0760, and AUC increases from 0.8607 to 0.9302. These gains indicate that the DDA descriptors improve not only prediction accuracy but also probability calibration and discriminative capability under cross-user generalization.

The observed improvements can be attributed to the structured, causally motivated design of the DDA descriptors, which encode timing alignment, headroom, and short-term traffic variability information derived from application-layer observables. By augmenting base-per-window statistics with descriptors explicitly aligned with XR system timing constraints, the model receives richer contextual information for QoE risk estimation. While Table~\ref{tab:dda_effect} does not isolate the contribution of individual descriptors, the aggregate performance gains demonstrate the effectiveness of the DDA formulation as a whole in improving robustness and calibration across users.

\begin{table}[ht]
\centering

\caption{Effect of DDA descriptors on QoE risk estimation performance under participant-independent evaluation (U1 train / U2 test).}
\scriptsize

\renewcommand{\arraystretch}{1.2}
\begin{adjustbox}{width=0.92\columnwidth}
\begin{tabular}{l c c c}
\hline
\textbf{Feature Set} & \textbf{SMAPE $\downarrow$} & \textbf{ECE $\downarrow$} & \textbf{AUC $\uparrow$} \\
\hline
Base per-window features & 0.2831 & 0.1399 & 0.8607 \\
DDA features          & \textbf{0.1524} & \textbf{0.0760} & \textbf{0.9302} \\
\hline
\end{tabular}
\end{adjustbox}
\label{tab:dda_effect}
\end{table}

\begin{table}[t]
\centering
\caption{\rev{Ablation study of DDA descriptor groups for QoE estimation.}}
\scriptsize
\renewcommand{\arraystretch}{1.2}
\begin{adjustbox}{width=\columnwidth}
\begin{tabular}{|l|c|c|c|}
\hline
\rev{\textbf{Feature Set}} & \rev{\textbf{QWK}} & \rev{\textbf{Macro-F1}} & \rev{\textbf{ECE}} \\
\hline
\rev{Base per-window descriptors} & \rev{0.7999} & \rev{0.5228} & \rev{0.1413} \\
\hline
\rev{Base + pacing-related descriptors} & \rev{0.7676} & \rev{0.4940} & \rev{0.1382} \\
\hline
\rev{Base + short-term stability descriptors} & \rev{0.7938} & \rev{0.4993} & \rev{0.1300} \\
\hline
\rev{Base + composite instability index} & \rev{\textbf{0.8219}} & \rev{\textbf{0.5242}} & \rev{\textbf{0.1196}} \\
\hline
\rev{Full DDA} & \rev{0.8064} & \rev{0.4972} & \rev{\textbf{0.1196}} \\
\hline
\end{tabular}
\end{adjustbox}
\label{tab:dda_ablation}
\end{table}

\rev{Table~\ref{tab:dda_ablation} reports an ablation study of the DDA descriptor groups. The base per-window descriptors provide the fundamental traffic state, while pacing-related descriptors characterize alignment with the inferred display refresh target. The short-term stability descriptors capture temporal variability and drift, and the composite instability index aggregates timing variability and pacing shortfall into a bounded QoE-sensitive descriptor. Among the evaluated feature groups, adding the composite instability index provides the largest gain, achieving the highest QWK and Macro-F1 and the lowest ECE. The full DDA representation also improves calibration compared with the base descriptor set, reducing ECE from 0.1413 to 0.1196 while maintaining strong agreement. These results indicate that the DDA performance is mainly driven by instability-aware temporal descriptors, which capture QoE-relevant traffic irregularities beyond base per-window traffic statistics.}

\subsection{Effect of Residual Learning}
\label{subsec:residual_significance}

\rev{For XR traffic prediction, residual learning improves short-term deviations superimposed on longer temporal trends. Tables~\ref{tab:p1}--\ref{tab:p4} show positive SMAPE gains for all backbone--task pairs, with the largest gain of 17.84\% for Transformer-based inter-arrival-time prediction and smaller recurrent-model gains in some settings. The residual pathway is therefore best interpreted as a lightweight causal refinement layer for structured prediction errors.}

\begin{table}[t]
\centering
\caption{\rev{Ablation of the residual bias term for frame-size prediction on Dataset~I. Results are reported as mean $\pm$ standard deviation over five independent runs.}}
\scriptsize
\renewcommand{\arraystretch}{1.15}
\begin{adjustbox}{width=\columnwidth}
\begin{tabular}{|l|c|c|c|}
\hline
\rev{\textbf{Residual configuration}} &
\rev{\textbf{RMSE}} &
\rev{\textbf{MAPE}} &
\rev{\textbf{SMAPE}} \\
\hline
\rev{Without bias ($b=0$)} &
\rev{$4908.67 \pm 42.31$} &
\rev{$0.0076 \pm 0.00008$} &
\rev{$0.97 \pm 0.009$} \\
\hline
\rev{With training-set bias} &
\rev{$4764.95 \pm 31.84$} &
\rev{$0.0054 \pm 0.00006$} &
\rev{$0.76 \pm 0.007$} \\
\hline
\end{tabular}
\end{adjustbox}
\label{tab:bias_ablation}
\end{table}

\rev{Table~\ref{tab:bias_ablation} shows that incorporating the training-set-derived residual bias improves frame-size prediction across all reported metrics. Relative to the zero-bias residual configuration, the training-set bias reduces RMSE from 4908.67 to 4764.95, corresponding to an approximate 2.9\% reduction. MAPE decreases from 0.0076 to 0.0054, and SMAPE decreases from 0.97 to 0.76, corresponding to approximate reductions of 28.9\% and 21.6\%, respectively. The experiments are repeated over five independent runs, and the smaller standard deviations for the biased configuration indicate more stable performance across random initializations. Because the bias is computed only once from the training residuals and then fixed during validation, testing, and inference, these gains reflect correction of a systematic residual offset rather than leakage from future test samples.}

\begin{figure}[t]
  \centering
  \subfloat[ResLearn-XR\label{fig:res_on}]
  {\includegraphics[width=0.48\linewidth]{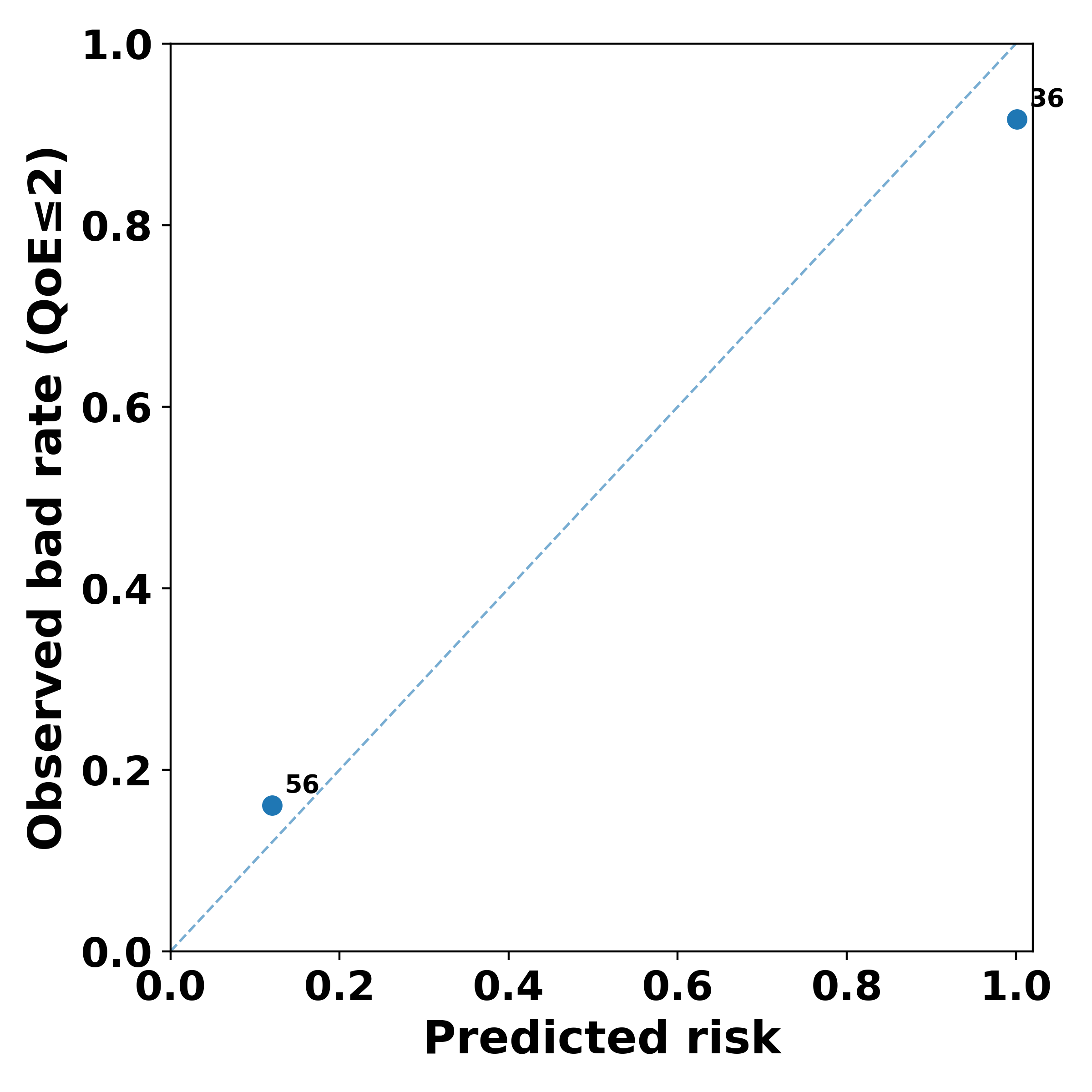}}\hfill
  \subfloat[Non-ResLearn-XR\label{fig:res_off}]
  {\includegraphics[width=0.48\linewidth]{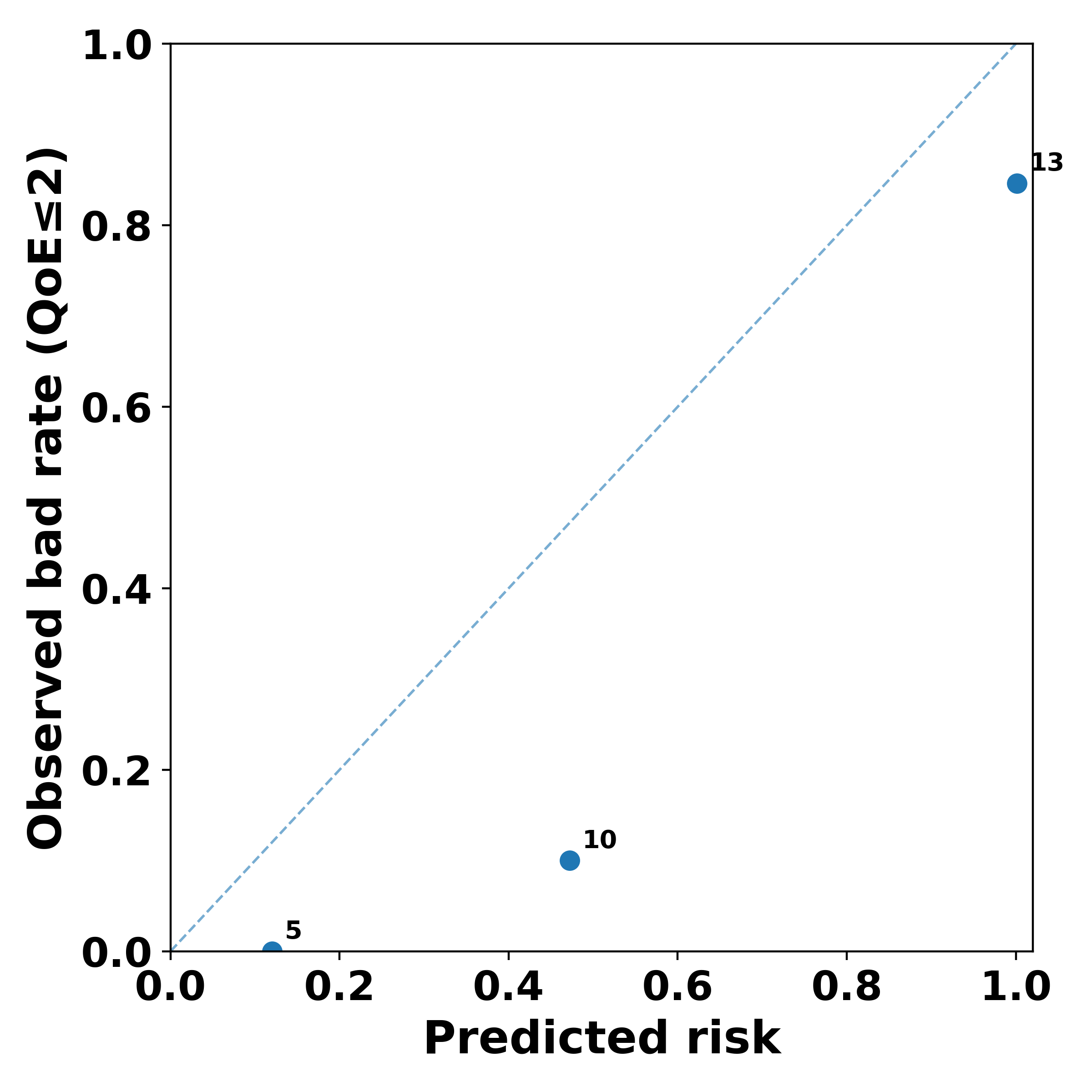}}

    \caption{Reliability comparison of QoE risk calibration on U2 test data. 
    For the displayed model pair, the ResLearn-XR model in (a) aligns more closely with the diagonal than the non-ResLearn-XR baseline in (b).}

  \label{fig:reliability_res}
\end{figure}

\rev{For QoE risk estimation, residual learning primarily improves predictive accuracy and discrimination, while calibration remains backbone- and split-dependent. The reliability diagrams in Fig.~\ref{fig:reliability_res} provide an illustrative example in which the displayed ResLearn-XR model is better calibrated than the displayed non-residual baseline; ECE trends across all models are reported in Tables~\ref{tab:qoe_u1u2}--\ref{tab:qoe_random}.}

\subsection{\rev{Deployment Feasibility}}
\label{subsec:deployment}

\rev{ResLearn-XR is intended as a lightweight online predictive component deployed at an edge server, access-network controller, or monitoring gateway. Its inputs are causally derived from packet timing, packet size, and packet direction, thus requiring no payload inspection or application-layer decryption. Table~\ref{tab:complexity} shows that the Transformer-based ResLearn-XR configuration adds only approximately 4K parameters, 0.02~MB of FP32 trainable-weight memory, and 0.05~ms per prediction window over the base Transformer (269K parameters, 1.08~MB FP32 trainable-weight memory, and 0.47~ms total), while the recurrent variants remain below 1~ms. These results support low-latency model-only inference, but they do not include packet capture, descriptor extraction, controller communication, scheduling, or actuation delay; therefore, full deployment still requires system-level profiling and closed-loop validation.}

\subsection{\rev{Limitations}}
\label{subsec:limitations}

\rev{The main limitations are the small two-participant QoE dataset, session-level weak labels assigned to all temporal windows, and the binary reduction of four ordinal subjective ratings. As a result, the participant-independent results should be interpreted as preliminary cross-user evidence rather than broad user-level generalization, and the QoE model may capture participant-, application-, bandwidth-, or session-specific correlates. Broader validation with larger cohorts, temporally localized QoE annotations, additional devices and network conditions, and controller-in-the-loop experiments is required before making broad deployment claims.}

\section{Conclusion and Future Work}

\rev{ResLearn-XR is a two-stage residual learning framework for cognitive XR network management, comprising value-space and logit-space residual heads for traffic prediction and QoE risk estimation, respectively. Grounded in ITU-T QoE causality via frame-timing-aware DDA descriptors, the corrected strictly causal traffic-prediction results show positive SMAPE gains up to 17.84\% and an average gain of approximately 6.7\% across the updated traffic tables, while QoE risk estimation achieves up to 87.8\% SMAPE improvement over single-stage baselines.}

\rev{The current QoE evaluation is limited by the two-participant dataset and session-level weak labels, so the participant-independent results should be interpreted as preliminary cross-user evidence rather than broad user-level generalization. ResLearn-XR should be viewed as a predictive modeling component for supporting proactive bandwidth allocation, scheduling, rate adaptation, admission control, and edge-resource orchestration, rather than as an end-to-end closed-loop controller. Future work will expand the XR Traffic-QoE dataset and evaluate heterogeneous HMDs and 5G/6G access conditions, while controller-in-the-loop deployment is left for future work.} 

\appendices

\section{XR Traffic Spikes and MTP Violation}
\label{appendix:mtp_relation}

The MTP latency decomposes as
$$
L_{\mathrm{MTP}}(t)=
L_{\mathrm{sense}}+L_{\mathrm{render}}+L_{\mathrm{enc}}+
L_{\mathrm{net}}(t)+L_{\mathrm{dec}}+L_{\mathrm{display}},
$$
where sensing, rendering, encoding, decoding, and display delays are near-constant relative to network dynamics. Grouping them as $L_{\mathrm{proc}}$, the ITU-T 20~ms constraint imposes a network latency budget $M = 20\,\mathrm{ms} - L_{\mathrm{proc}}$, and MTP stability requires $L_{\mathrm{net}}(t)\le M$.

Over a window of duration $\Delta_W \approx W\bar{\iota}_t$, the total offered traffic is $A(t)=c_t s_t$ (bits), giving arrival rate
$$
\lambda(t)=\frac{c_t s_t}{\Delta_W}\approx\frac{s_t}{\iota_t} \quad \text{(bits/s)},
$$
since $c_t \approx \Delta_W/\iota_t$, and utilization $\rho(t)=\lambda(t)/C < 1$.
This shows that all three traffic metrics $(s_t, c_t, \iota_t)$ jointly determine network load.

Modeling the bottleneck as an M/M/1 queue~\cite{shortle2018fundamentals} with service rate $C$, the expected queueing delay per bit is $\mathbb{E}[W_q^{(\mathrm{bit})}(t)] = \rho(t)/[C(1-\rho(t))]$, so the frame-level network delay is
$$
L_{\mathrm{net}}(t)\approx L_0 + \frac{s_t}{C-\lambda(t)},
$$
where $L_0$ accounts for fixed propagation delays.
As $\rho(t)\to 1^{-}$, differentiation yields
$$
\frac{\partial\,\mathbb{E}[W_q^{(\mathrm{bit})}]}{\partial\rho(t)}=\frac{1}{C(1-\rho(t))^2},
$$
showing that even small burst-driven increases in $s_t$ or $c_t$, or reductions in $\iota_t$, can exhaust $M$ and violate the 20~ms constraint. For general traffic, Kingman's approximation for a G/G/1 queue gives
$$
\mathbb{E}[W_q^{(\mathrm{bit})}(t)] \approx
\frac{\rho(t)}{1-\rho(t)}\frac{c_a^2 + c_s^2}{2C},
$$
where $c_a^2$ and $c_s^2$ are the squared coefficients of variation of inter-arrival and service times. The same $(1-\rho)^{-1}$ divergence persists, confirming that bursty XR traffic produces disproportionate latency increases and motivating explicit residual correction of short-horizon predictions of $(s_t, c_t, \iota_t)$.

\rev{This relationship also motivates the proxy-based MTP-risk analysis in Section~\ref{subsec:traffic_results}, where each prediction window uses
\begin{equation}
\widehat{L}_{k}=\frac{\widehat{c}_{k}\widehat{s}_{k}}{\widehat{\iota}_{k}+\epsilon},
\end{equation}
with $\epsilon$ preventing division by zero. A burst-risk indicator is then
\begin{equation}
\widehat{v}_{k}=\mathbb{I}(\widehat{L}_{k}>C_{\mathrm{eff}}),
\end{equation}
where $C_{\mathrm{eff}}$ is an effective service-capacity threshold. Residual correction reduces burst underestimation and improves proxy-risk detection, while direct closed-loop MTP-violation measurement remains future work.}

\section{\rev{Implementation and Reproducibility Configuration}}
\label{appendix:repro_config}

\begin{table}[H]
\centering
\caption{\rev{Implementation and reproducibility configuration for ResLearn-XR.}}
\footnotesize
\renewcommand{\arraystretch}{0.88}
\begin{adjustbox}{width=\columnwidth}
\begin{tabular}{|p{0.32\columnwidth}|p{0.62\columnwidth}|}
\hline
\rev{\textbf{Item}} & \rev{\textbf{Configuration}} \\
\hline

\multicolumn{2}{|c|}{\rev{\textbf{Traffic prediction branch}}} \\
\hline
\rev{Targets} &
\rev{FIA-derived frame count, average frame size, and average frame inter-arrival time.} \\
\hline
\rev{Window duration $W$} &
\rev{Dataset I: $1$~s; Datasets II--III: $0.5$~s.} \\
\hline
\rev{Rolling aggregation window} &
\rev{Frame size/count/IAT windows: Dataset I $20/22/5$; Dataset II Exp. 1 $20/22/10$; Dataset II Exp. 2--6 $20/22/5$; Dataset III Exp. 1--2 $25/20/5$.} \\
\hline
\rev{Base look-back $S$} &
\rev{Dataset I: $17$; Dataset II Exp. 1: $20$; Dataset II Exp. 2--6: $5$; Dataset III Exp. 1--2: $8$ windows.} \\
\hline
\rev{Residual look-back $S_r$} &
\rev{$S_r=1$; only the latest observed bias-shifted residual $\tilde{e}_k$ is used to predict $\hat{r}_{k+1}$.} \\
\hline
\rev{Training split} &
\rev{Chronological within-session $40/10/50\%$ train/validation/test split; sequences are constructed within each segment to avoid split-boundary leakage.} \\
\hline
\rev{Base traffic loss} &
\rev{MSE between $\hat{y}^{(\mathrm{base})}_{k+1}$ and $y_{k+1}$.} \\
\hline
\rev{Residual traffic loss} &
\rev{MSE between $g_{\phi}^{(\mathrm{traf})}(\tilde{e}_k)$ and $e_{k+1}$.} \\
\hline
\rev{Optimizer} &
\rev{Adam in TensorFlow/Keras; optimizer states are excluded from Table~\ref{tab:complexity} FP32 memory.} \\
\hline
\rev{Epochs and batch size} &
\rev{$20$ epochs; batch size $1$.} \\
\hline
\rev{Weight decay} &
\rev{None in the released traffic-prediction configuration.} \\
\hline
\rev{Random seed} &
\rev{Fixed where specified; multi-run tables report mean $\pm$ standard deviation.} \\
\hline

\multicolumn{2}{|c|}{\rev{\textbf{Traffic prediction architectures}}} \\
\hline
\rev{Backbones} &
\rev{Transformer, LSTM, GRU, and Stacked LSTM under the two-stage residual framework in Table~\ref{tab:complexity}.} \\
\hline
\rev{Transformer} &
\rev{Encoder with multi-head self-attention, feed-forward projection, dropout, and dense output head.} \\
\hline
\rev{LSTM / GRU} &
\rev{Single recurrent backbone and dense output head.} \\
\hline
\rev{Stacked LSTM} &
\rev{Two-layer recurrent backbone with inter-layer dropout and dense output head.} \\
\hline
\rev{Residual traffic head} &
\rev{Lightweight dense value-space correction head trained on the corrected causal residual target.} \\
\hline
\rev{Complexity accounting} &
\rev{Table~\ref{tab:complexity} counts trainable parameters for the active encoder, prediction head, and residual head; FP32 memory uses four bytes/weight and excludes activations, buffers, and optimizer states.} \\
\hline
\rev{Inference timing} &
\rev{Single-window, batch-size-one latency after descriptor construction, compared with the 20~ms MTP budget; packet capture, control, scheduling, and actuation delays require separate profiling.} \\
\hline

\multicolumn{2}{|c|}{\rev{\textbf{QoE risk-estimation branch}}} \\
\hline
\rev{Input descriptors} &
\rev{Causal DDA history $\mathbf{H}^{(S)}_k=[\mathbf{h}_{k-S+1},\ldots,\mathbf{h}_k]$.} \\
\hline
\rev{DDA look-back $U$} &
\rev{Fixed causal horizon for $\mathrm{CoV}_U(\cdot)$ and $\beta_U(\cdot)$: $U=16$.} \\
\hline
\rev{QoE label} &
\rev{Binary weak-supervision label from session-level QoE annotation.} \\
\hline
\rev{Class weight} &
\rev{$\beta=N_0/N_1$, using non-elevated-risk ($N_0$) and elevated-risk ($N_1$) training samples.} \\
\hline
\rev{QoE base loss} &
\rev{Class-weighted binary cross-entropy with logits.} \\
\hline
\rev{QoE residual head} &
\rev{Lightweight dense logit-space correction head applied before probability calibration.} \\
\hline
\rev{QoE residual loss} &
\rev{Class-weighted binary cross-entropy on corrected logit $\ell_k^{(\mathrm{base})}+\delta_k$.} \\
\hline
\rev{Calibration split} &
\rev{Dedicated split disjoint from base and residual training data.} \\
\hline
\rev{Calibration method} &
\rev{Stratum-wise isotonic regression using DDA-derived covariates only.} \\
\hline
\rev{Calibration strata} &
\rev{Inferred refresh anchor $\hat r$, offered-load bin $\mathcal{B}_r(r_k)$, pacing-ratio bin $\mathcal{B}_{\rho}(\rho_k)$, and instability bin $\mathcal{B}_J(J_k)$.} \\
\hline
\rev{Calibration back-off} &
\rev{Use the most specific non-empty stratum calibrator; otherwise use the global isotonic calibrator.} \\
\hline

\end{tabular}
\end{adjustbox}
\label{tab:repro_config}
\end{table}

\bibliographystyle{IEEEtran}
\bibliography{refs}

\makeatletter
\def\@textbottom{\vskip \z@ \@plus 100fil}
\makeatother

\begin{IEEEbiography}
    [{\includegraphics[width=1in,height=1.25in,clip,keepaspectratio]{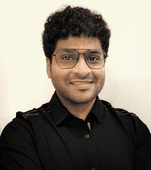}}]{Yoga Suhas Kuruba Manjunath}
    (Member, IEEE) received the Bachelor of Engineering degree in electronics and communication from Visvesvaraya Technological University, Belagavi, India, in 2014, the Master of Engineering degree in artificial intelligence from Toronto Metropolitan University (TMU) (formerly Ryerson University), Toronto, ON, Canada, in 2021, and the Ph.D. degree from the Department of Electrical and Computer Engineering, TMU, in 2025.

He has over five years of industry experience as an Internet-of-Things (IoT) architect, developing IoT stacks for the dairy and hospitality industries that have impacted over two hundred thousand customers. His combined hardware and software expertise has contributed to numerous projects in these sectors. Upon returning to academia, he focused on advancing his research skills, resulting in publications in leading conferences and journals, including IEEE GLOBECOM, IEEE WF-IoT, and \textit{Electronic Commerce Research and Applications}. He actively participates in communications-related projects at the Communications and Signal Processing Applications Laboratory and Ubiquitous Intelligent Communication and Computing at TMU. His current research interests include AI-based IoT solutions and virtual-reality network optimization for quality of service.

Dr. Manjunath received the Best Team Award at the IEEE Leaders of Tomorrow event organized by IEEE Toronto. He serves as a peer reviewer for several journals, including \textit{IEEE Transactions on Wireless Communications}, \textit{IEEE Transactions on Network and Service Management}, \textit{IEEE Internet of Things Journal}, \textit{IEEE Open Journal of the Communications Society}, \textit{Electronic Commerce Research and Applications}, and \textit{The Journal of Supercomputing}. He serves as vice chair of the IEEE Vehicular Technology Chapter of the IEEE Toronto Section and is a member of the IEEE Vehicular Technology Society and the IEEE Communications Society.
\end{IEEEbiography}

\begin{IEEEbiography}
    [{\includegraphics[width=1in,height=1.25in,clip,keepaspectratio]{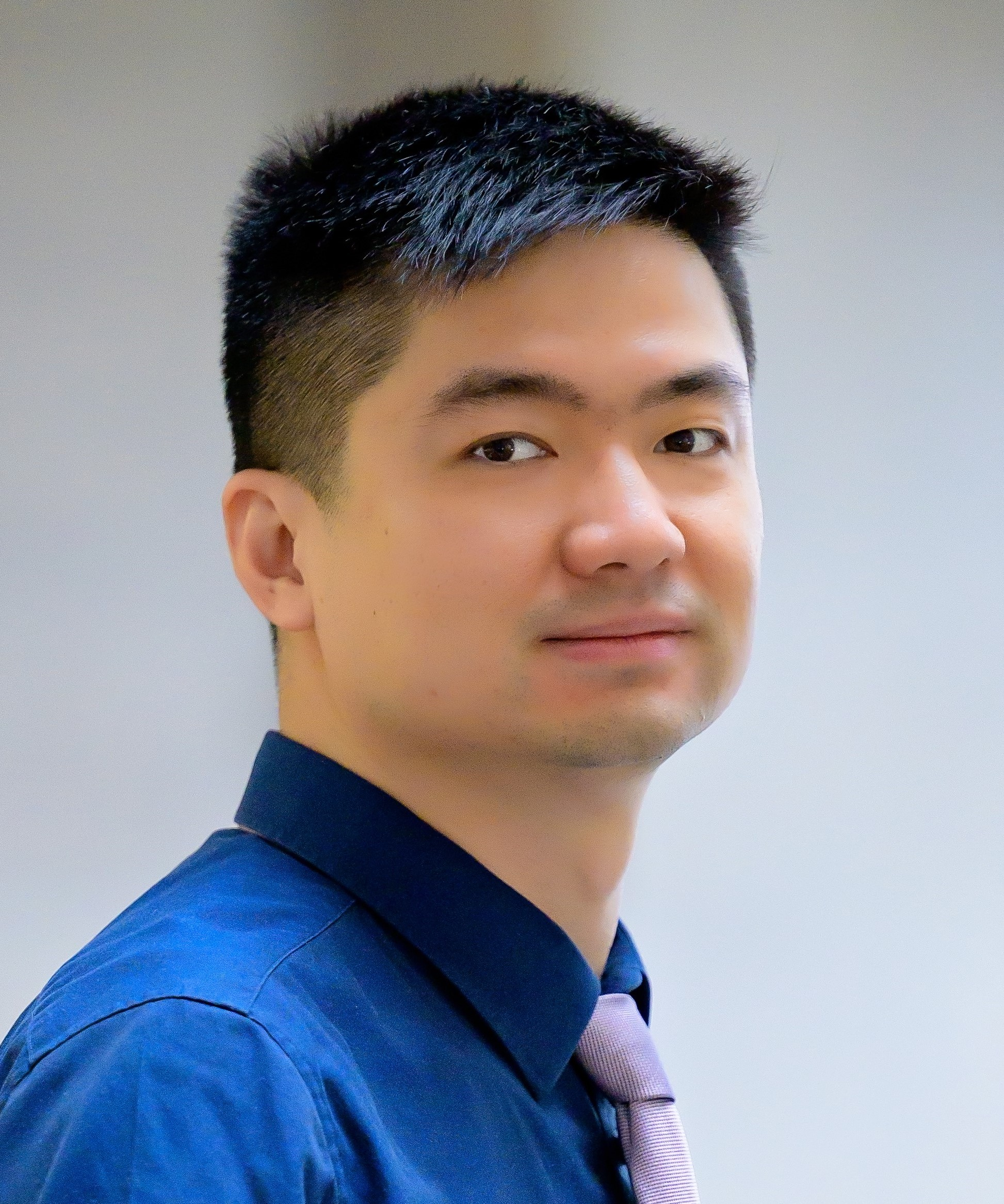}}]{Jie Gao}
    (Senior Member, IEEE) is an Assistant Professor with the School of Information Technology, Carleton University, Ottawa, ON, Canada. His research interests include machine learning for communications and networking, XR and immersive communications, and emerging network technologies for 6G. He is serving or has served as an Editor for \textit{IEEE Transactions on Cognitive Communications and Networking}, \textit{IEEE Open Journal of the Communications Society}, and the Vehicular Technology Section of \textit{IEEE Access}. He has co-chaired symposia, tracks, and workshops at IEEE conferences, including IEEE GLOBECOM, VTC, ICCC, and INFOCOM. He received the IEEE Vehicular Technology Society Open Journal of Vehicular Technology Best Paper Award in 2025, the IEEE Best Land Transportation Paper Award in 2024, and the Wisconsin Space Grant Consortium Early-Stage Investigator Grant Award in 2021.
\end{IEEEbiography}

\begin{IEEEbiography}
    [{\includegraphics[width=1in,height=1.25in,clip,keepaspectratio]{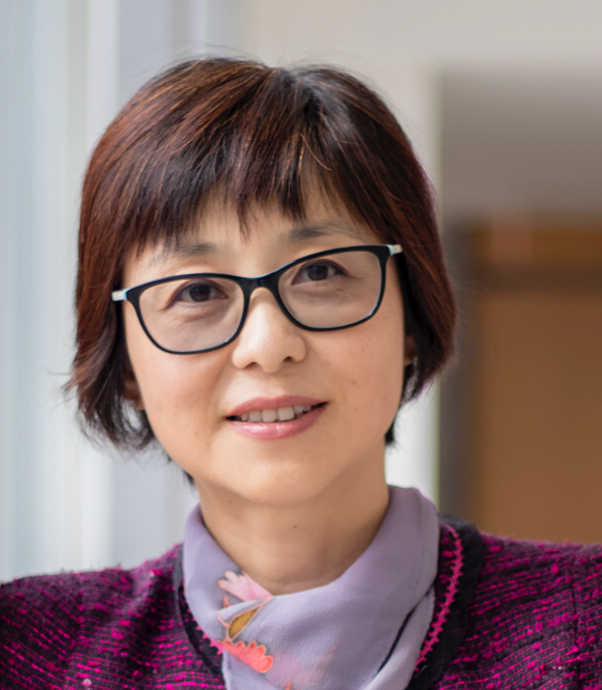}}]{Lian Zhao}
    (Fellow, IEEE) received the Ph.D. degree from the Department of Electrical and Computer Engineering (ELCE), University of Waterloo, Canada, in 2002. She joined the Department of Electrical and Computer Engineering at Toronto Metropolitan University (formerly Ryerson University), Canada, in 2003. Her research interests are in the areas of wireless communications, resource management, mobile edge computing, IoT/IoV networks, and machine learning for communications.

    She has been an IEEE Communication Society (ComSoc) and IEEE Vehicular Technology (VTS) Distinguished Lecturer (DL); received the Best Land Transportation Paper Award from IEEE Vehicular Technology Society in 2016 and 2024, Best Paper Award from the 2013 International Conference on Wireless Communications and Signal Processing (WCSP), and the Canada Foundation for Innovation (CFI) New Opportunity Research Award in 2005.
    
    She has been serving as an Editor for IEEE Transactions on Wireless Communications, IEEE Internet of Things Journal, and IEEE Transactions on Vehicular Technology (2013-2021). She serves as a TPC Chair for VTC2025-Fall, a co-Chair of Wireless Communication Symposium for IEEE Globecom 2020/2025, and IEEE ICC 2018; Finance co-Chair for 2021 ICASSP; Local Arrangement co-Chair for IEEE VTC Fall 2017 and IEEE Infocom 2014. She has been an elected member for the Board of Governor (BoG) of VTS since 2023.  
\end{IEEEbiography}

\end{document}